%% file: main.tex
\documentclass[runningheads]{llncs}

\let\llncssubparagraph\subparagraph
\let\subparagraph\paragraph
\usepackage[compact]{titlesec}
\let\subparagraph\llncssubparagraph

\usepackage{graphicx}

\usepackage{tikz}
\usepackage{comment}
\usepackage{amsmath,amssymb} 
\usepackage{color}

\usepackage{amsmath}
\usepackage{amssymb}
\usepackage{booktabs}
\usepackage[export]{adjustbox}
\usepackage{multirow}
\usepackage{adjustbox}
\usepackage{colortbl}
\usepackage{color}
\usepackage{subcaption}
\usepackage{graphicx} 
\definecolor{mygray}{RGB}{220,220,220}
\usepackage{xcolor}
\usepackage{xspace}
\usepackage[pagebackref,breaklinks,citecolor=blue,colorlinks,linkcolor=red,bookmarks=false]{hyperref}
\usepackage{siunitx}
\usepackage{booktabs}
\usepackage{blindtext}
\usepackage{hyperref}

\usepackage{mathtools, nccmath}

\usepackage{caption}
\DeclareCaptionFont{9pt}{\fontsize{9pt}{10pt}\selectfont}

\usepackage[capitalize]{cleveref}
\usepackage[T1]{fontenc}
\usepackage[font=small,labelfont=bf,tableposition=top]{caption}

\DeclareCaptionLabelFormat{andtable}{#1~#2  \&  \tablename~\thetable}

\makeatletter
\DeclareRobustCommand\onedot{\futurelet\@let@token\@onedot}
\def\@onedot{\ifx\@let@token.\else.\null\fi\xspace}

\def\eg{\emph{e.g}\onedot}

\makeatother

\crefname{section}{Sec.}{Secs.}
\Crefname{section}{Section}{Sections}
\Crefname{table}{Table}{Tables}
\crefname{table}{Tab.}{Tabs.}

\def\ourmetric{TETA\xspace}

\newcommand{\parsection}[1]{\vspace{1mm}\noindent\textbf{#1 }}

\usepackage[accsupp]{axessibility}  

\usepackage{titlesec}

\begin{document}
\titlespacing*{\section}{0pt}{0.28cm}{0.28cm}
\titlespacing*{\subsection}{0pt}{0.2cm}{0.2cm}

\pagestyle{headings}
\mainmatter
\def\ECCVSubNumber{4933}  


\title{Tracking Every Thing in the Wild} 

\titlerunning{Tracking Every Thing in the Wild} 
\author{Siyuan Li, Martin Danelljan, Henghui Ding, Thomas E. Huang, Fisher Yu}
\institute{Computer Vision Lab, ETH Z{\"u}rich\\
\url{http://vis.xyz/pub/tet}}

\maketitle

\begin{abstract}

Current multi-category Multiple Object Tracking (MOT) metrics use class labels to group tracking results for per-class evaluation. Similarly, MOT methods typically only associate objects with the same class predictions. 
These two prevalent strategies in MOT implicitly assume that the classification performance is near-perfect. 
However, this is far from the case in recent large-scale MOT datasets, which contain large numbers of classes with many rare or semantically similar categories. Therefore, the resulting inaccurate classification leads to sub-optimal tracking and inadequate benchmarking of trackers.
We address these issues by disentangling classification from tracking.
We introduce a new metric, Track Every Thing Accuracy (TETA), breaking tracking measurement into three sub-factors: localization, association, and classification, allowing comprehensive benchmarking of tracking performance even under inaccurate classification. TETA also deals with the challenging incomplete annotation problem in large-scale tracking datasets. We further introduce a Track Every Thing tracker (TETer), that performs association using Class Exemplar Matching (CEM). Our experiments show that TETA evaluates trackers more comprehensively, and TETer achieves significant improvements on the challenging large-scale datasets BDD100K and TAO compared to the state-of-the-art.
\end{abstract}

\input{introduction.tex}
\input{relatedworks.tex}
\input{method.tex}
\input{experiments.tex}
\input{appendix}

\clearpage
%
%
\bibliographystyle{splncs04}
\bibliography{egbib}

\end{document}

%% file: introduction.tex
\section{Introduction}

Multiple Object Tracking (MOT) aims to estimate the trajectory of objects in a video sequence.
While common MOT benchmarks \cite{kitti,mot17,mot20}  only consider tracking objects from very few pre-defined categories, \eg, pedestrian and car, the number of categories of interest in the real world is overwhelming. 
Although the recent extension of MOT to a large number of categories \cite{bdd100k,tao} may seem trivial, it raises profound questions about the definition and formulation of the problem itself, which are yet to be addressed by the community.

\begin{figure}[t]
	\centering
	\includegraphics[width=\linewidth]{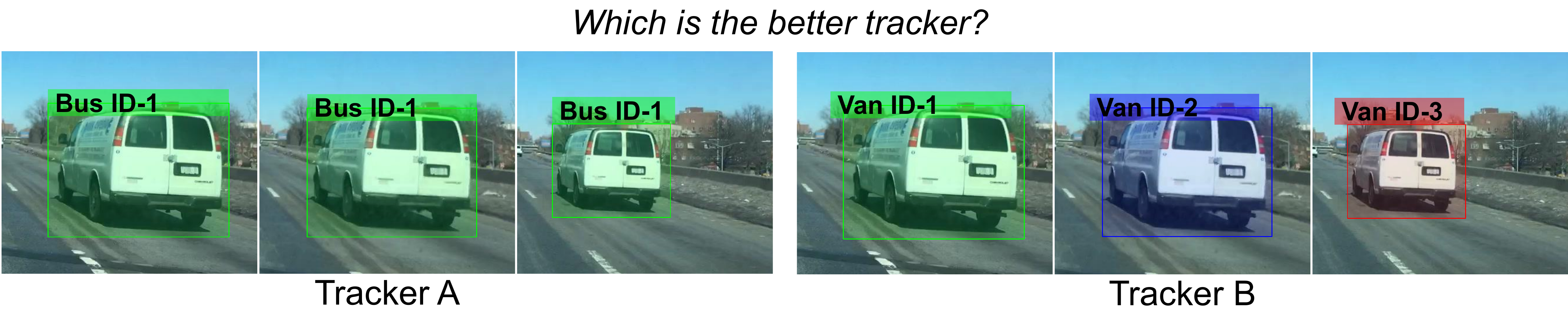}\vspace{-0.2cm}
	\caption{Tracking results from two different trackers (A and B). The same color means the same track. Tracker A gets 0 score in terms of the MOTA~\cite{MOTA}, IDF1~\cite{IDF1}, and HOTA~\cite{hota} metrics, while the tracker B gets 33 for first two and 44 for HOTA 
	}
	\label{fig:teaser}
\end{figure}

In~\figurename~\ref{fig:teaser}, we show tracking results from two different trackers on the same video sequence.
Tracker A tracks the object perfectly, but with a slightly incorrect classification on a fine-grained level.
Tracker B classifies the object perfectly but does not track the object at all. 
\emph{Which one is the better tracker?}
The mMOTA~\cite{MOTA} metric gives a 0 score for tracker A and a score of 33 for tracker B.
The above example raises an interesting question: \textit{\textbf{Is tracking still meaningful if the class prediction is wrong?}}
In many cases, the trajectories of wrongly classified or even unknown objects are still valuable.
For instance, an autonomous vehicle may occasionally track a van as a bus, but the estimated trajectory can equally well be used for path planning and collision avoidance.

Current MOT models and metrics~\cite{bergmann2019tracking,zhou2020tracking,MOTA,IDF1,russakovsky2015imagenet,hota} are mainly designed for single-category multiple object tracking. When extending MOT to the large-scale multi-category scenarios, they simply adopt the same single-category metrics and models by treating each class independently.
The models first detect and classify each object, and then the association is only done between objects of the same class.
Similarly, the metrics use class labels to group tracking results and evaluate each class separately.
This implicitly assumes that the classification is good enough since it is the prerequisite for conducting association and evaluating tracking performance.

The aforementioned near-perfect classification accuracy is mostly valid on benchmarks consisting of only a handful of common categories, such as humans and cars.
However, it does not hold when MOT extends to a large number of categories with many rare or semantically similar classes.
The classification itself becomes a very challenging task on imbalanced large-scale datasets such as LVIS~\cite{lvis}.
Also, it is difficult to distinguish similar fine-grained classes because of the naturally existing class hierarchy, \eg, the bus and van in~\figurename~\ref{fig:teaser}.
Besides, many objects do not belong to any predefined category in real-world settings. Thus, treating every class independently without accounting for the inaccuracy in classification leads to inadequate benchmarking and non-desired tracking behavior.
To expand tracking to a more general scenario, we propose that classification should be disentangled from tracking, in both evaluation and model design, for multi-category MOT. 
To achieve this, we design a new metric, Track Every Thing Accuracy (TETA), and a new model, Track Every Thing tracker (TETer). 

The proposed TETA metric disentangles classification performance from tracking.
Instead of using the predicted class labels to group per-class tracking results, we use location with the help of local cluster evaluation.
We treat each ground truth bounding box of the target class as the anchor of each cluster and group prediction results inside each cluster to evaluate the localization and association performance.
Our local clusters enable us to evaluate tracks even when the class prediction is wrong.
Furthermore, the local cluster evaluation makes TETA competent to deal with incomplete annotations, which are common in datasets with a large number of classes, such as TAO~\cite{tao}.

Our TETer follows an \textit{Associate-Every-Thing (AET)} strategy.
Instead of associating objects in the same class, we associate every object in neighboring frames. 
The AET strategy frees association from the challenging classification/detection issue under large-scale long-tailed settings. However, despite wholly disregarding the class information during association, we propose a new way of leveraging it, which is robust to classification errors. We introduce Class Exemplar Matching (CEM), where the learned class exemplars incorporate valuable class information in a soft manner. In this way, we effectively exploit semantic supervision on large-scale detection datasets while not relying on the often incorrect classification output. CEM can be seamlessly incorporated into existing MOT methods and consistently improve performance. Moreover, our tracking strategy enables us to correct the per-frame class predictions using rich temporal information.

\begin{figure}[!tb]
    \centering
        \includegraphics[width=1\linewidth]{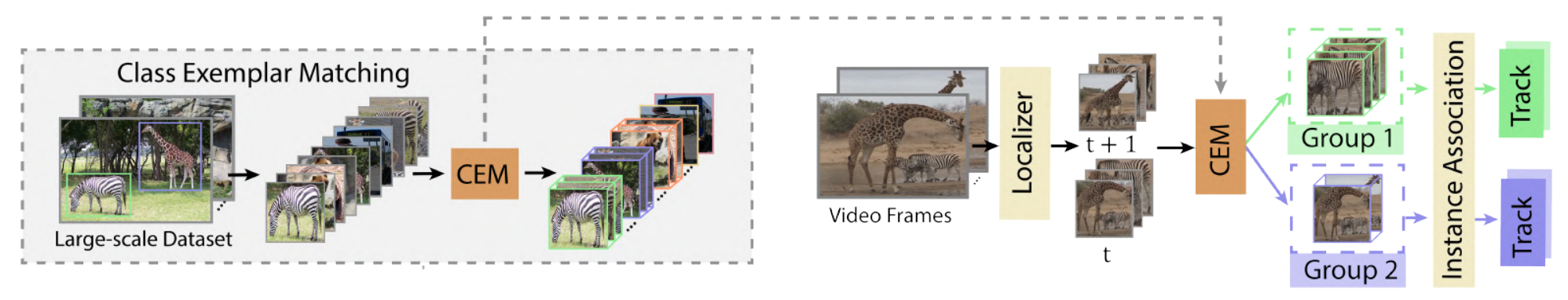}
        \vspace{-0.4cm}
            \caption{CEM can be trained with large-scale datasets and directly employed for tracking}
	\label{fig:train_pipeline}
\end{figure}

We analyze our methods on the newly introduced large-scale multi-category tracking datasets, TAO~\cite{tao} and BDD100K~\cite{bdd100k}.
Our comprehensive analysis show that our metric evaluate trackers more comprehensively and achieve better cross dataset consistency despite incomplete annotations.
Moreover, our tracker achieves state-of-the-art performance on TAO and BDD100K, both when using previously established metrics and the proposed TETA. 

%% file: relatedworks.tex
\section{Related Work}
\parsection{Multi-Object Tracking (MOT)}aims to track multiple objects in video sequences.
Earlier methods follow a track-first paradigm, which do not rely on classification during tracking~\cite{trackdecide,rcnnproposal,athar2020stem}.
Some utilize LiDAR data with model-free detection~\cite{kaestner2012generative,moosmann2013joint,dewan2016motion} or point cloud segmentation~\cite{teichman2012tracking,teichman2011towards}.
Others~\cite{trackdecide,mitzel2012taking,ovsep2016multi} first segment the scene~\cite{sceneseg}, which enables tracking of generic objects.
Recently, the most common paradigm for MOT is tracking-by-detection, focusing on learning better appearance features to strengthen association~\cite{leal2016learning,milan2017online,yang2019video,lu2020retinatrack,meinhardt2021trackformer,liu2020gsm}, modeling the displacement of each tracked object~\cite{bergmann2019tracking,zhou2020tracking,peng2020chained}, or using a graph-based approach~\cite{schulter2017deep,braso2020learning}. Previous MOT approaches mostly focus on benchmarks with a few common categories, while recent works \cite{openingup,dave2019towards} study the MOT in open-set settings where the goal is to track and segment any objects regardless of their categories. Those methods use a class agnostic trained detector or RPN network to generate object proposals, while classification is essential in many applications, \eg, video analysis. The close-set settings with large-scale, long-tailed datasets are severely under-explored. We study the MOT in such a scenario, identifying issues and proposing solutions in both model design and evaluation metric.

\parsection{MOT Metrics} often evaluate both detection and association performance. Multi-Object Tracking Accuracy (MOTA)~\cite{MOTA} was first introduced to unify the two measures. MOTA performs matching on the detection level and measures association performance by counting the number of identity switches. IDF1~\cite{IDF1} and Track-mAP instead performs matching on the trajectory level. Recently, Higher-Order Tracking Accuracy (HOTA)~\cite{hota} was proposed to fairly balance both components by computing a separate score for each. Liu et al.~\cite{openingup} proposes a recall-based evaluation to extend MOT into open-world settings. All above metrics do not independently access the classification performance, making them unsuitable for large-scale multi-category MOT. TETA extends HOTA by further breaking down detection into localization and classification, enabling TETA to evaluate association despite classification failures. Furthermore, current metrics have issues when evaluating trackers on non-exhaustively annotated datasets such as TAO, which TETA can handle. 

%% file: method.tex
\section{Tracking-Every-Thing Metric}
Here we introduce the Track Every Thing Accuracy (\ourmetric) metric. 
We first discuss how classification is handled in current metrics and the incomplete annotation problem in \sectionautorefname~\ref{sec:incomplete}.
Then, we formulate \ourmetric in \sectionautorefname~\ref{sec:teta} to address the existing issues.
\subsection{Limitations for Large-scale MOT Evaluation}

\parsection{How to handle classification.}
\label{sec:overemphasize}
How to evaluate classification in MOT is an important but under-explored problem. MOT metrics such as MOTA~\cite{MOTA}, IDF1~\cite{IDF1}, and HOTA~\cite{hota} are designed for the single category MOT. When extending to multiple classes, they require trackers to predict a class label for each object, then they group tracking results based on the labels and evaluate each class separately. However, the wrong classification happens frequently in long-tailed scenarios, which leads to failures in the grouping based on class labels, and the tracking performance will not be evaluated even if the tracker localizes and tracks the object perfectly as shown in ~\figurename~\ref{fig:teaser}. \par
One simple solution is to ignore classification and evaluate every object class-agnostically. However, large vocabulary datasets often follow long-tailed distributions where few classes dominate the dataset. 
Ignoring class information leads to the evaluation being dominated by those classes, resulting in trackers' performance in tracking rare classes being negligible.
The class-aware HOTA proposes to use a geometric mean between with classification confidence and HOTA, which requires the trackers to output the class probability distribution, while most only output the final categories. Moreover, it still cannot access classification independently.

\begin{figure}[!tb]
    
    \begin{minipage}[t]{.5\textwidth}
    \centering
 \includegraphics[trim=200 10 150 250, clip,  height=0.4\linewidth]{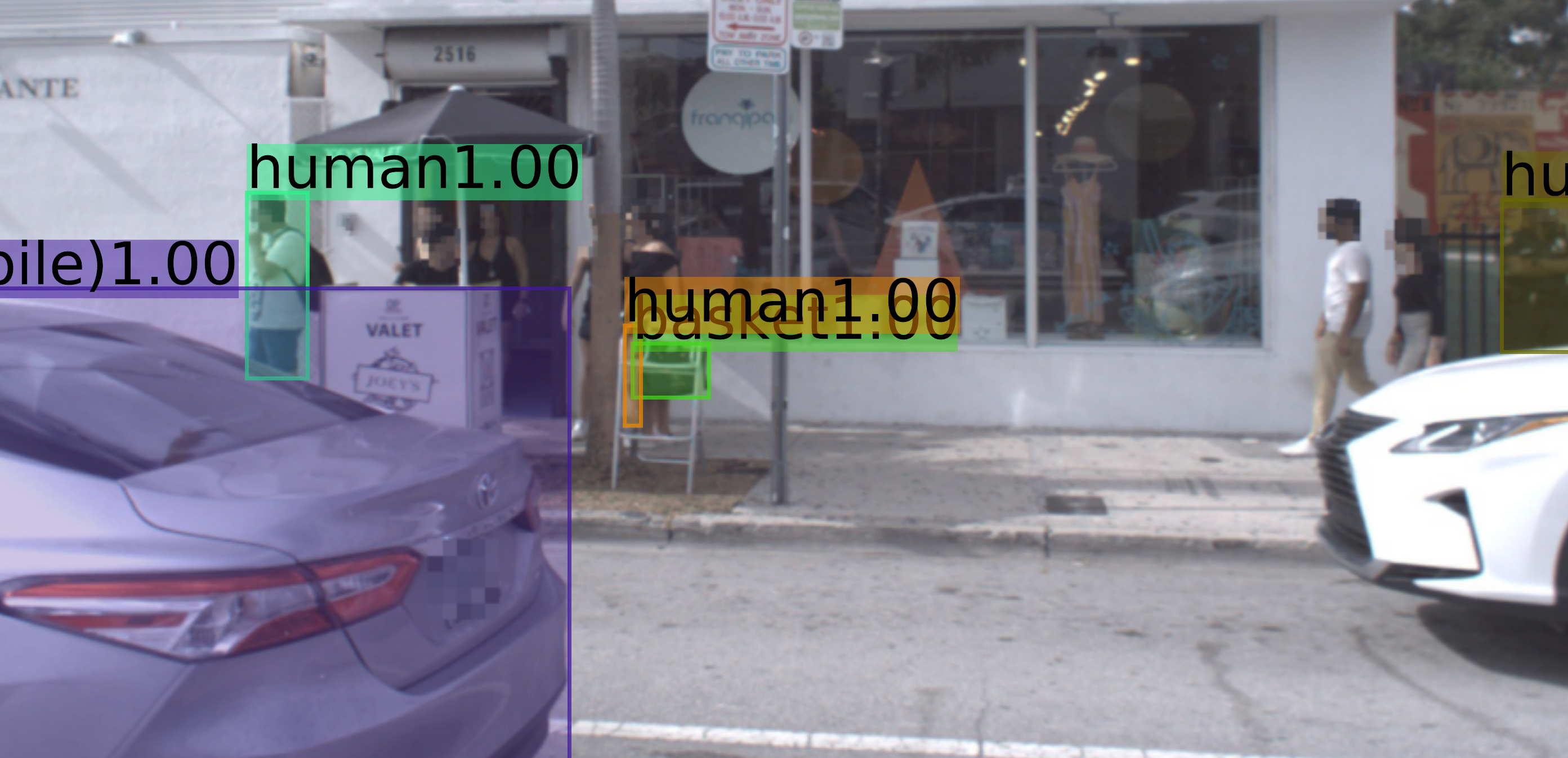}
    \end{minipage}%
    \begin{minipage}[t]{0.5\textwidth}
        \centering
\includegraphics[trim=250 10 80 250, clip, height=0.4\linewidth]{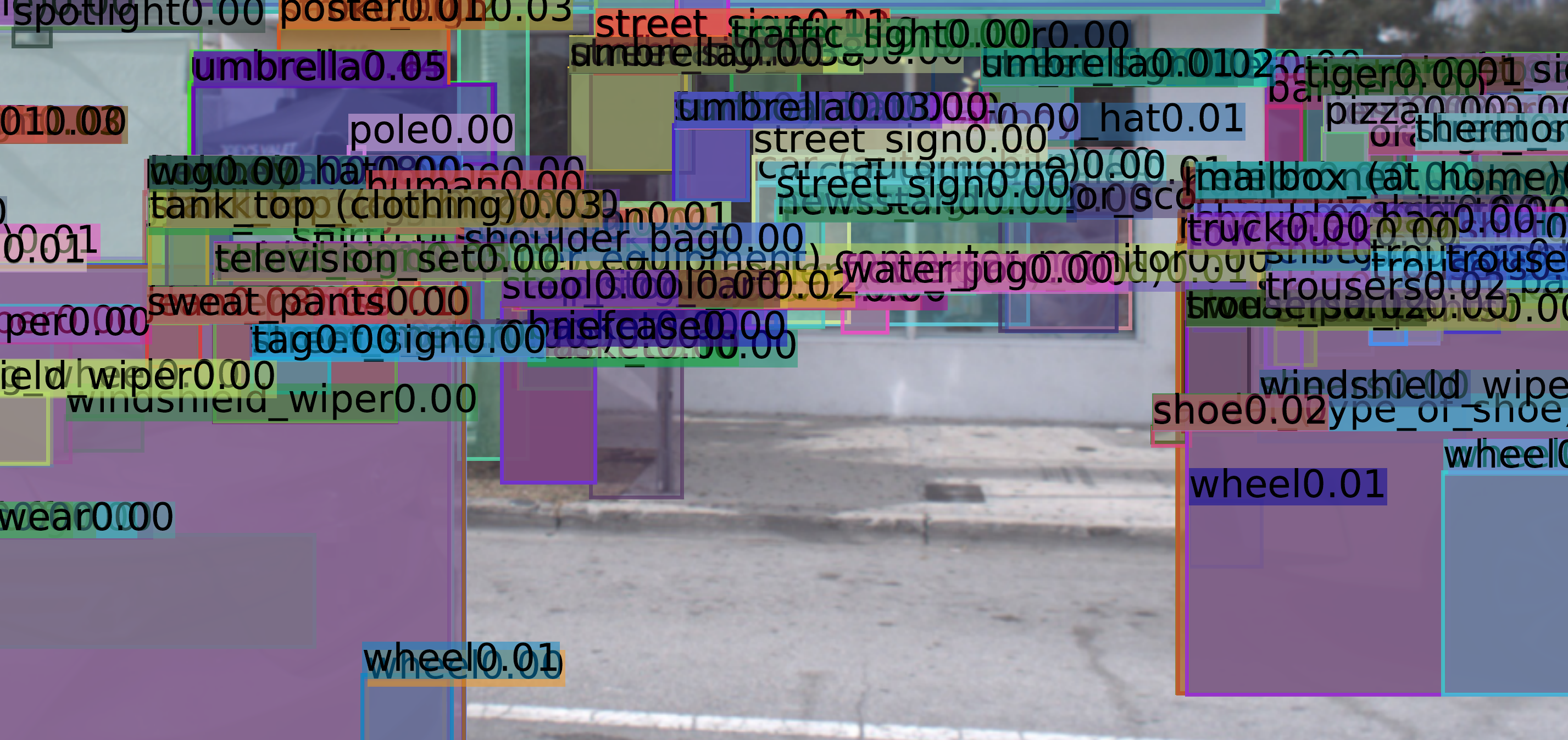} 
        \label{fig:prob1_6_1}
    \end{minipage}
    \vspace{-0.3cm}
	\caption{\textbf{Left}: TAO ground truth sample. TAO is partially annotated. \textbf{Right}: Corresponding prediction from the best tracker AOA~\cite{aoa} ranked by TAO metric. AOA generates many low confidence bounding boxes, making it difficult to use in practice }
	\label{fig:tao_gt}
\end{figure}

\parsection{Incomplete Annotations.}
\label{sec:incomplete}
MOT metrics such as MOTA~\cite{MOTA}, IDF1~\cite{IDF1}, and HOTA~\cite{hota} are designed for datasets with exhaustive annotations of every object.
However, it is prohibitively expensive to annotate every object when constructing large-scale datasets with many categories.
The TAO dataset contains over 800 categories, but most of them are not exhaustively annotated (see~\figurename~\ref{fig:tao_gt}).
Incomplete annotations pose a new challenge: \textit{how can we identify and penalize false positive (FP) predictions?}
MOTA, IDF1, and HOTA metrics treat every unmatched prediction as FP, but this falsely penalizes correct predictions with no corresponding annotations.
On the other hand, TAO metric~\cite{tao} adopts the same federated evaluation strategy as the LVIS~\cite{lvis} dataset and does not penalize categories if there is no ground-truth information about their presence or absence.
This strategy inadvertently rewards a large number of false positives.
In~\figurename~\ref{fig:tao_gt}, we visualize predictions from the best tracker on TAO.
Since TAO does not punish most false positives, trackers are incentivized to generate many low confidence tracks to increase the chances that objects from rare categories get tracked, making their results difficult to be used in practice.
Also, this makes TAO a game-able metric.
We show a simple copying and pasting trick that can drastically improve the TAO metric's score in \sectionautorefname~\ref{sec:trick_tao}.
A similar issue is also observed in the LVIS mAP metric~\cite{dave2021evaluating}.

\subsection{Tracking-Every-Thing Accuracy (\ourmetric)}
\label{sec:teta}
\ourmetric builds upon the HOTA~\cite{hota} metric, while extending it to better deal with multiple categories and incomplete annotations. \ourmetric consists of three parts: a localization score, an association score, and a classification score, which enable us to evaluate the different aspects of each tracker properly. 

\begin{figure}[!tb]
    \centering
    \begin{minipage}{.45\textwidth}
        \includegraphics[width=0.95\linewidth]{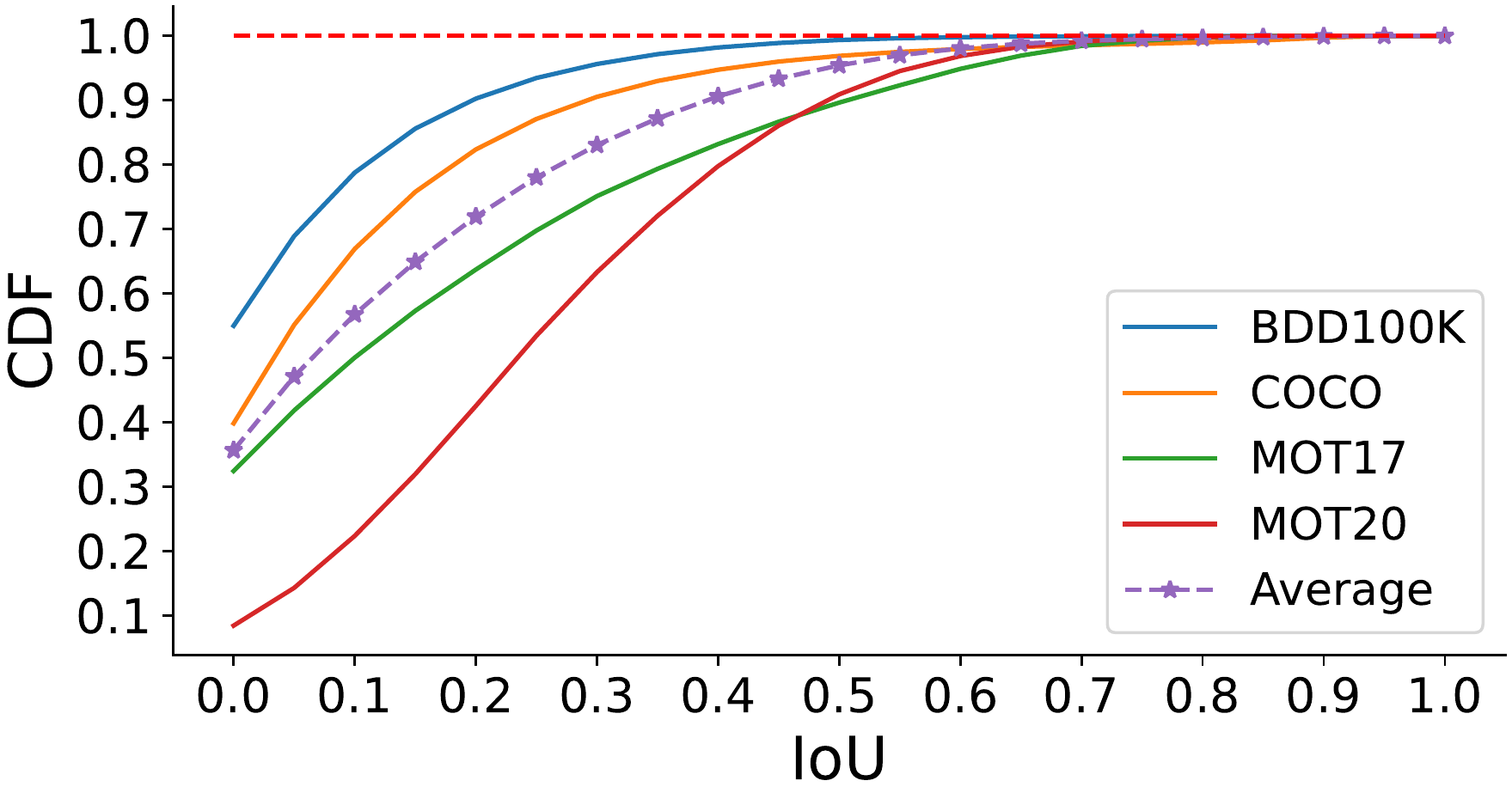}
    \end{minipage}%
    \begin{minipage}{0.55\textwidth}
        \centering
        \includegraphics[width=0.85\linewidth]{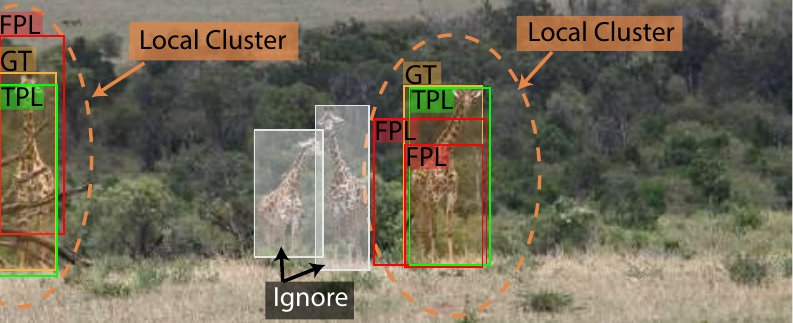}
        \label{fig:prob1_6_1}
    \end{minipage}
    \vspace{-0.3cm}
	\caption{\textbf{Left}: Inter-object overlap in real datasets. We compute the cumulative probability of ground truth bounding boxes that have different level of IoU overlaps in four different datasets with exhaustive annotations along with their average. Extreme inter-object overlap is very rare in real data. \textbf{Right}: Local cluster evaluation. $\mathrm{TPL}$, $\mathrm{FPL}$, and $\mathrm{GT}$ are the true positive localization, false positive localization, and ground truth, respectively. We create a cluster for each ground truth bounding box based on the IoU similarities. For evaluation, we only consider predictions inside each cluster. The predictions that do not belong to any cluster will be ignored}
	\label{fig:local-cluster}
\end{figure}

\parsection{Local cluster evaluation.}
We design the local clusters to deal with incomplete annotations and disentangle classification from large-scale tracking evaluation. The main challenge for evaluation with incomplete annotations is determining false positives. We propose local cluster evaluation to strike a balance between false-penalizing or the non-penalizing phenomenon as discussed in ~\ref{sec:incomplete}. We have observed that even though we do not have exhaustive annotations, we can still identify a specific type of false positives with high confidence.
Unlike previous metrics, we only consider predictions within local clusters. We view each ground truth bounding box as an anchor point of a cluster and assign each prediction to the closest anchor points within an IoU margin of $r$. The predictions inside the clusters not chosen as matched true positives are considered false positives. \figurename~\ref{fig:local-cluster} shows the inter-object overlap in popular object detection and tracking datasets, which indicates extreme inter-object overlap is rare in the real world. If we set the $r$ to 0.7 or higher, there is less than 1\% chances that we make mistakes even in highly crowded dataset like MOT20. 

To avoid false punishments, we ignore the predictions that are not assigned to any clusters during evaluation. This process is illustrated in \figurename~\ref{fig:local-cluster}. The margin $r$ of the clusters can be set according to different scenarios. The bigger the $r$ is, the more conservative the metric is regarding choosing false positives. Also, it means fewer false punishments. If the dataset is super crowded and lacks annotation,  we can select a higher $r$ to avoid false punishment.
The local cluster design also allows us to disentangle classification. For evaluation of a particular class, we evaluate predictions that are assigned to clusters with ground truth bounding boxes of that class. Since the per class result grouping is done using location instead of classification. Thus, within each local cluster, we are able to evaluate the tracking performance even if the class predictions are wrong.

\parsection{Localization Score.}
The localization score measures the localization performance of a tracker.
A true positive candidate $b \in\mathrm{TPL}$ is a prediction box (pBox) that has an IoU higher than a localization threshold $\alpha$ with a ground truth box (gBox).
We use the Hungarian algorithm~\cite{hungarian} to choose the final matched $\mathrm{TPL}$ that optimizes both localization and association scores.
The chosen assignment represents the best-localized tracks.
A false negative localization ($\mathrm{FNL}$) is a gBox that is not matched to any other pBox.
A false positive localization ($\mathrm{FPL}$) is defined based on each cluster.
If a pBox is in a cluster but is not matched to any ground truth, it is a false positive. 
The localization score is computed using the Jaccard Index,
\begin{equation}
   \mathrm{LocA} = \frac{|\mathrm{TPL}|}{|\mathrm{TPL}| + |\mathrm{FPL}| + |\mathrm{FNL}|}.
\end{equation}

\parsection{Association Score.}
Our association score follows the definition of HOTA but redefines true positive associations (TPA), false negative associations (FNA), and false positive associations (FPA) to be based on each $b$.
The association score of $b$ is\useshortskip
\begin{equation}
  \mathrm{AssocA}(b) = \frac{|\mathrm{TPA}(b)|}{|\mathrm{TPA}(b)| + |\mathrm{FPA}(b)| + |\mathrm{FNA}(b)|}.
\end{equation}
The final association score is the average over all TPLs,
\begin{equation} 
  \mathrm{AssocA} = \frac{1}{|\mathrm{TPL}|}\sum_{b \in \mathrm{TPL}}\mathrm{AssocA}(b).
\end{equation}

\parsection{Classification Score.}
The classification score reflects the pure performance of the classifier in each tracker.
Unlike all other tracking metrics where classification performance is entangled with tracking performance, our metric makes it an independent score.
We only consider the well-matched TPL, where $\alpha$ is at least 0.5.
The classification score is defined for each class.
The true positive classification ($\mathrm{TPC}$) for certain class $c$ is defined as
\begin{equation}
\mathrm{TPC}(c) = \{b| b \in \mathrm{TPL} ~ \wedge~ \mathrm{pc}(b) = \mathrm{gc}(b) = c \},
\end{equation}
where $\mathrm{pc}(b)$ is the class ID of $b$ and $\mathrm{gc}(b)$ is the class ID of the ground truth that is matched to $b$.
This set includes all TPL that have the same predicted class as the corresponding ground truth.
The false negative classification for class $c$ is defined as
\begin{equation}
\mathrm{FNC}(c) = \{b| b \in \mathrm{TPL} ~ \wedge~ \mathrm{pc}(b) \neq c \wedge \mathrm{gc}(b) = c \},
\end{equation}
which includes all TPL that have incorrect class predictions with ground truth class $c$.
The false positive classification for class $c$ is defined as
\begin{equation}
\mathrm{FPC}(c) = \{b| b \in \mathrm{TPL} ~ \wedge~ \mathrm{pc}(b) = c \wedge \mathrm{gc}(b) \neq c \}
\end{equation}
which includes all TPL with class $c$ but is matched to an incorrect ground truth class. If the dataset is fully annotated, the $b \in \mathrm{P}$, which includes TPL and the predictions outside clusters. Full annotations indicate that the predictions that are far away from gBox wrongly classify background or other classes as $c$.
The final classification score is
\begin{equation}  
   \mathrm{ClsA} = \frac{|\mathrm{TPC}|}{|\mathrm{TPC}| + |\mathrm{FPC}| + |\mathrm{FNC}|}.
\end{equation}

\parsection{Combined Score.}
HOTA uses geometric mean to balance detection and association. However, the geometric mean becomes zero if any term is zero.
If the classification performance of a tracker is close to zero, \eg due to a long-tail class distribution, it will completely dominate the final tracking metric if computed as a geometric mean. Therefore, we use an arithmetic mean to compute the final score:
 \begin{equation}
   \mathrm{TETA} = \frac{\mathrm{LocA} + \mathrm{AssocA} + \mathrm{ClsA}}{3}.
\end{equation}
Besides, since different applications focus on different aspects, we encourage users to look at each subfactor based on the needs instead of focusing on a single score.

\section{Tracking-Every-Thing Tracker}

\begin{figure*}[tb]
	\centering
    \includegraphics[width=\linewidth]{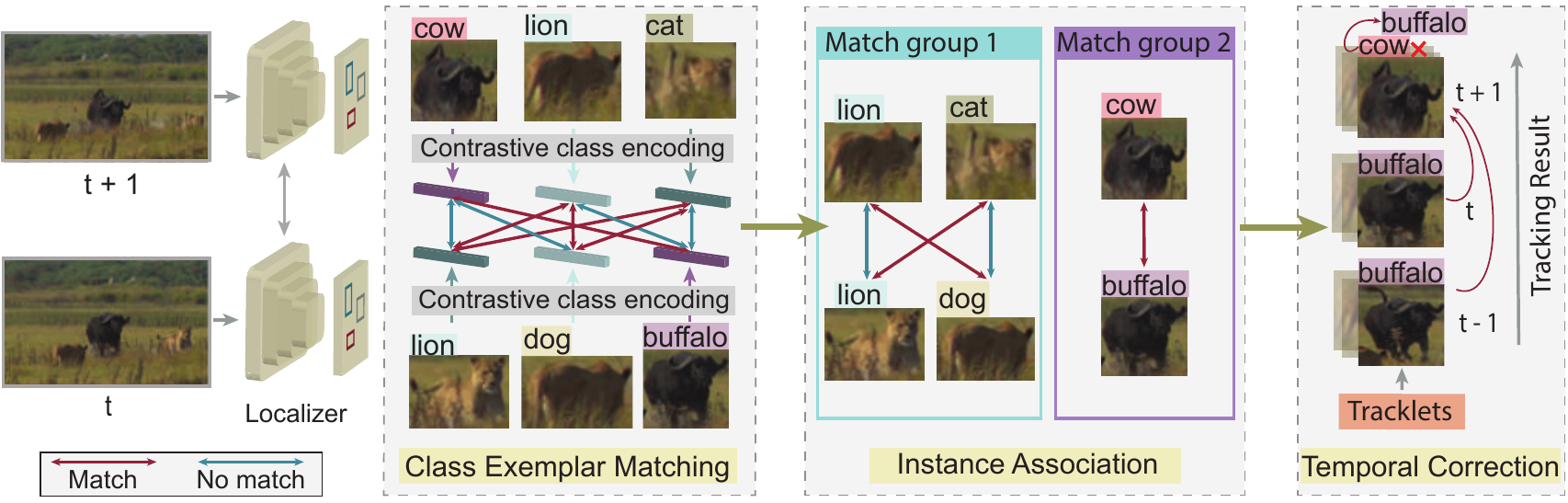} 
	\caption{ Association protocol of TETer. For every pair of frames, we first compute and match the class exemplars for each localized object to determine potential matching candidates. Then, we perform instance association to determine the final matches. Finally, we use rich temporal information to correct classification errors in each frame}
	\label{fig:pipeline}
\end{figure*}

We here introduce our Tracking Every Thing tracker (TETer). TETer aims to discover every foreground object, associate, and classify them across time. The full pipeline is shown in~\figurename~\ref{fig:pipeline}.

\subsection{Class-Agnostic Localization (CAL)}

To track every thing, we first need to localize them.
However, object detectors struggle on large-scale, long-tailed datasets, especially for rare categories.
Interestingly, when decoupling localization and classification in common object detectors, we find that the detector can still localize rare or even novel objects well.
\figurename~\ref{fig:faster-rcnn} shows a comparison of the object detector's performance with and without considering classification on the TAO validation set.
When we do not consider class predictions during evaluation, the performance of the detector is stable across rare, common, and frequent classes.
This strongly suggests that the bottleneck in detection performance lies in the classifier.
With this in mind, we replace the commonly used intra-class non-maximum suppression (NMS) using class confidence with a class-agnostic counterpart to better localize every object in the wild.
\subsection{Associating Every Thing}

Association is often done by considering a single or a combination of cues, \eg, location, appearance, and class. Learning motion priors under large-scale, long-tailed settings is challenging since the motion patterns are irregular among classes.
Moreover, there are also many objects in the real world that are not in any predefined categories.
In contrast, objects in different categories usually have very different appearances.
Thus, we adopt appearance similarity as our primary cue. 

We propose an alternative approach for utilizing class information as feature clues during association. Instead of trusting the class predictions from object detectors and using them as hard prior, we learn class exemplars by directly contrasting samples from different categories.
This enables us to compute similarity scores for pairs of objects at the category-level, which can better deal with semantically similar classes compared to discrete class labels.
During association, we use the class exemplars to determine potential matching candidates for each object.
This process can be thought of as using class information as soft prior. Therefore, it can integrate fine-grained cues required for classification (\eg the difference between a big red bus and a red truck), which are difficult to learn for the purely class-agnostic appearance embedding.

\begin{figure}[!tb]
    
    \begin{minipage}[t]{.45\textwidth}
	\includegraphics[width=1\linewidth]{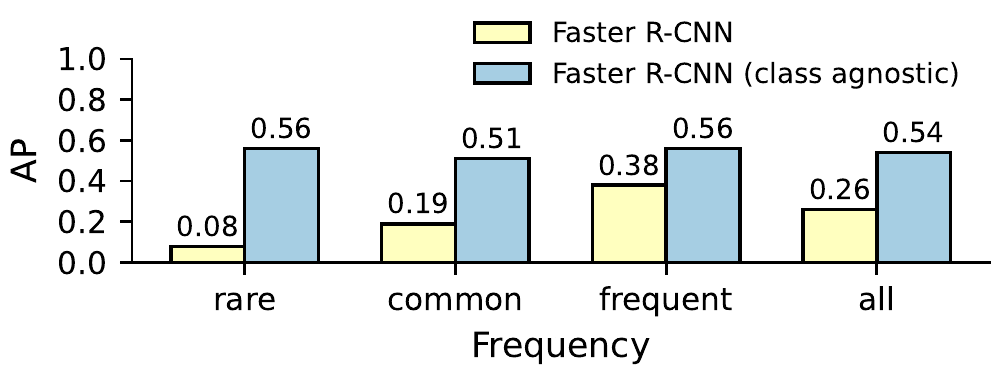}
	\caption{Faster R-CNN's performance w/ and w/o considering classification}
	\label{fig:faster-rcnn}
    \end{minipage}\hspace{0.7cm}
    \begin{minipage}[t]{0.5\textwidth}
	\includegraphics[width=1\linewidth]{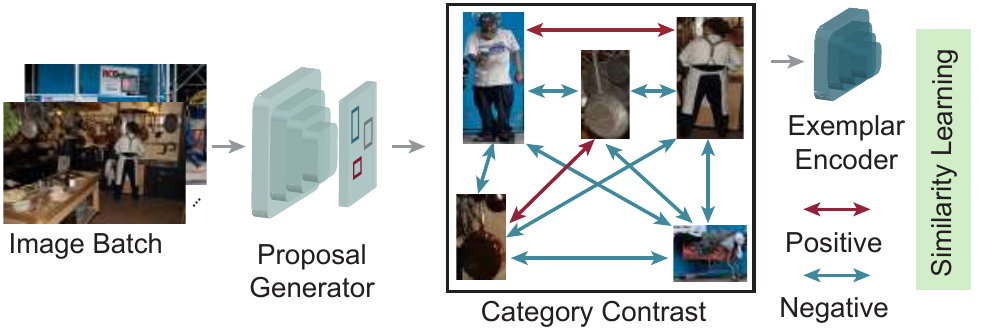}
	\caption{Training pipeline of CEM}
	\label{fig:training_pipeline}
    \end{minipage}
   \label{fig:game_tao}
\end{figure}
\parsection{Class Exemplar Matching (CEM).}~The training pipeline based on a two-stage detector is shown in ~\figurename~\ref{fig:training_pipeline}. The Region Proposal Network (RPN) computes all the Region of Interest (RoI) proposals from the input images.
Then, we use RoI align to extract the feature maps from the multi-scale feature outputs.
The feature maps are used as input for the exemplar encoder to learn category similarity.
The exemplar encoder generates class exemplars for each RoI.
We assign category labels to each RoI with a localization threshold $\alpha$.
If a RoI has an IoU higher than $\alpha$ (in our case, $\alpha = 0.7$) with a ground truth box, then we assign the corresponding category label to the RoI.
Positive samples are RoIs from the same category, and negative samples are those from different categories.

We adapt the SupCon loss~\cite{khosla2020supervised} and propose an unbiased supervised contrastive loss (U-SupCon):

\begin{equation}
\resizebox{0.9\hsize}{!}{$\mathcal{L_{C}}  = -\sum_{q \in Q} \frac{1}{|Q^+(q)|}\sum_{q^+ \in Q^+(q)}\mathrm{log}\frac{\mathrm{exp}(\mathrm{sim}(q, q^+)/ \tau)}{\mathrm{PosD}(q)+ \sum_{q^- \in Q^-}\mathrm{exp}(\mathrm{sim}(q, q^-)/ \tau)}$},
\end{equation}

\begin{equation}
\resizebox{0.6\hsize}{!}{$\mathrm{PosD}(q) = \frac{1}{\left | Q^+(q)) \right |}\sum_{q^+ \in Q^+}\mathrm{exp}(\mathrm{sim}(q, q^+)/\tau)$},
\end{equation}

where $Q$ is the set of class generated from a random sampled image batch, $Q^+(q)$ is the set of all positive samples to $q$, $Q^-(q)$ is the set of all negative samples to $q$, $\mathrm{sim(\cdot)}$ denotes cosine similarity, and $\tau$ is a temperature parameter. We set $\tau$ to 0.07. We add the $\mathrm{PosD(q)}$ to prevent the varying lower bound of the SupCon loss when training with detection proposals, where the number of positive samples is consistently changing. 

\parsection{Association Strategy.}
\label{sec:asso}
For a query object $q$ in a tracklet, we find a group of candidates by comparing their class exemplars.
Specifically, assume we have the class exemplar $q_c$ for the query object $q$ in frame $t$, and a set of detected objects $D$ in frame $t+1$ and their class exemplars $d_c \in D_c$.
We compute the similarities between $q_c$ and $D_c$ and select candidates with high similarity.
This gives us a candidate list $C=\{d\;|\;\mathrm{sim}(q_c, d_c) > \mathrm{\delta },\;d \in D \}$. $\mathrm{\delta }$ is set to 0.5. \par
To determine the final match from the candidate list, any existing association method can be used.
Thus, CEM can be readily used to replace existing hard prior-based matching.
For our final model TETer,
we further utilize quasi-dense similarity learning~\cite{qdtrack} to learn instance features for instance-level association. 
We compute instance-level matching scores with each candidate from $C$ using bidirectional softmax and cosine similarity. We take the candidate with the maximum score and if the score is larger than $\beta$, then it is a successful match. We set $\beta$ to 0.5.

\parsection{Temporal Class Correction (TCC).}
The AET strategy allows us to correct the classification using rich temporal information. If we track an object, we assume the class label to be consistent across the entire track. We use a simple majority vote to correct the per-frame class predictions. 

%% file: experiments.tex
\section{Experiments}

We conduct analyses of different evaluation metrics and investigate the effectiveness of our new tracking method on TAO~\cite{tao} and BDD100K~\cite{bdd100k}.
TAO provides videos and tracking labels for both common and rare objects with over 800 object classes. Although BDD100K for driving scenes has fewer labeled categories, some, like trains, are much less frequent than common objects such as cars. In this section, we first compare different metrics with \ourmetric. 
Then we evaluate the proposed TETer on different datasets and plug CEM into existing tracking methods to demonstrate its generalizability.

\noindent\textbf{Implementation Details.}~For the object detector, we use Faster R-CNN~\cite{ren2015faster} with Feature Pyramid Network (FPN)~\cite{Lin2017FeaturePN}. We use ResNet-101 as backbone on TAO, same as TAO baseline~\cite{tao}, and use ResNet-50 as backone on BDD100K, same as the QDTrack~\cite{qdtrack}.
On TAO, we train our model on a combination of LVISv0.5~\cite{lvis} and the COCO dataset using repeat factor sampling. The repeat factor is set to 0.001. We use the SGD optimizer with a learning rate of 0.02 and adopt the step policy for learning rate decay, momentum, and weight decay are set to 0.9 and 0.0001. We train our model for 24 epochs in total and learning rate is decreased at 16 and 22 epoch. For TETer with SwinT backbone~\cite{liu2021swin}, we use 3x schedule used by mmdetection~\cite{chen2019mmdetection}. For TETer-HTC, we use the HTC-X101-MS-DCN detector from ~\cite{li2020overcoming}. On BDD100K, we load the same object detector weights from QDTrack~\cite{qdtrack} and fine-tune the exemplar encoder with on BDD100K Detection dataset with other weights frozen. For each image, we sample a maximum of 256 object proposals. 
For more details, please refer to the appendix~\sectionautorefname~\ref{implement-details}.

\begin{figure}[!tb]
  
    \begin{minipage}{.55\textwidth}
        \includegraphics[width=0.97\linewidth]{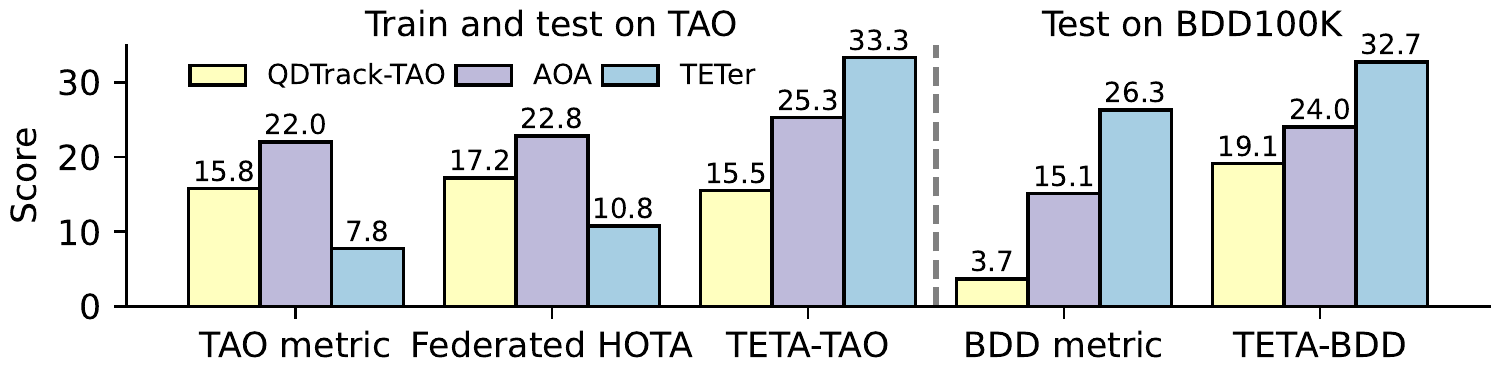}
    \end{minipage}%
    \begin{minipage}{0.45\textwidth}
        \centering
        \includegraphics[width=0.97\linewidth]{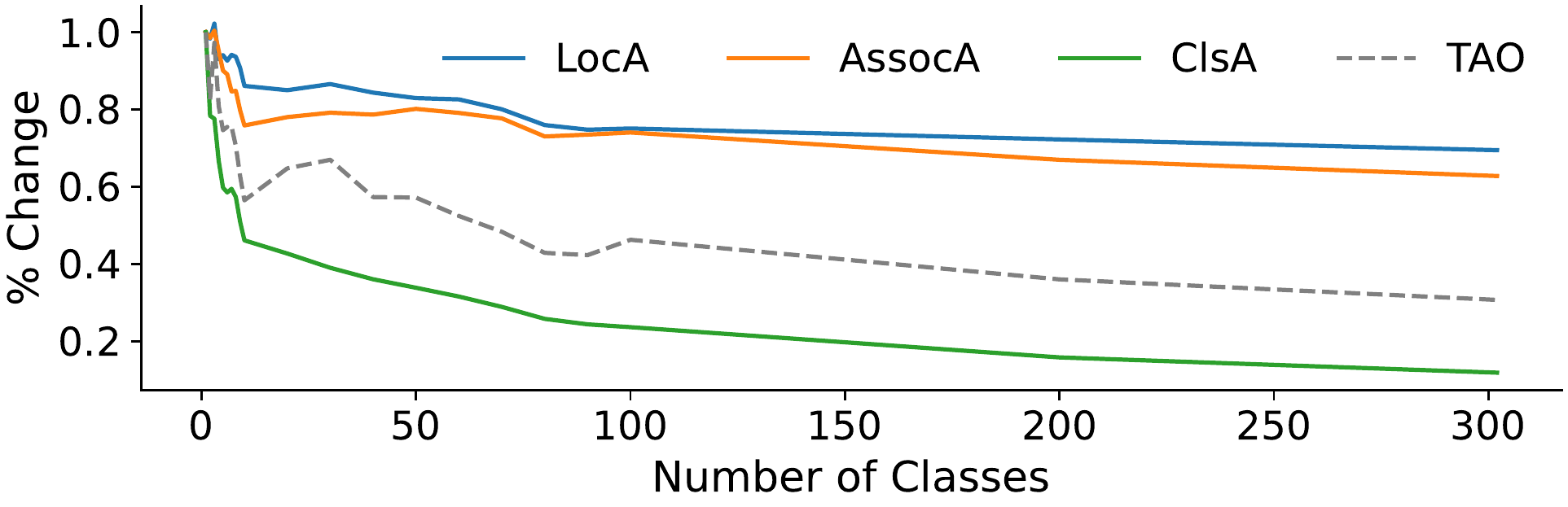}
    \end{minipage}
            \caption{\textbf{Left}: We pre-train models on TAO (incomplete annotations) and directly test them on BDD (complete annotations) with the default BDD metric (IDF1). We omit MOTA as its value range is $(-\infty, 1]$, which is inconsistent with other metrics. \textbf{Right}: Percentage change in score of each metric as number of evaluation classes increases}
	\label{fig:cross_dataset}
\end{figure}
\subsection{Metric Analysis}

\parsection{Cross Dataset Consistency.} A good metric should correlate with real world tracking performance.
Although we face difficulties in incomplete annotations, this principle for metric design should not change.
For instance, a tracker designed for tracking objects belonging to hundreds of categories on TAO should also work well on new video sequences that contain a subset of those categories.
We evaluate this by using the BDD100K dataset,
which has seven out of eight of its categories overlapped with TAO.  
We treat BDD100K as the new video sequences in the real world to test two trackers: QDTrack-TAO~\cite{qdtrack}, which is optimized for the TAO metric, and our tracker, which is optimized for \ourmetric.
We only evaluate on the overlapped categories, which also contains exhaustive annotations for every object.

As shown in~\figurename~\ref{fig:cross_dataset} (Left), the tracker selected by the TAO metric overly optimizes for the incomplete TAO dataset setting, which does not generalize well to BDD100K.
In comparison, the tracker selected by \ourmetric generalizes well to BDD100K.
Our metric gives the same ranking on the complete annotations setting with the default BDD100K IDF1 metric despite facing the difficulties of ranking trackers under incomplete annotations.

\parsection{Comprehensively Analyze Trackers.}
Correctly understanding different aspects of trackers is crucial for designing trackers for various scenarios.
For example, it is important for an autonomous vehicle to detect and understand the trajectory of every object to avoid collision, but slightly wrong classification may not be as critical.
In this experiment, we evaluate the effect of the number of classes on metric scores on the TAO validation set.
We use the same tracking predictions but merge the class predictions by sorting the classes in descending order based on the number of instances and combining the last $n$ classes.
For example for $n=2$, we merge all classes besides humans (the most frequent class) into a single class and only evaluate on two classes.
We sample several $n$ between 1 (single class) and 302 (all classes) and evaluate on each set of classes.

The result is shown in~\figurename~\ref{fig:cross_dataset} (Right).
Although the trajectory predictions are the same, the score produced by the TAO metric drops significantly as the number of classes increases. 
As the TAO metric entangles classification and tracking into a single metric, this makes it hard to determine which part of the tracker went wrong.
On the other hand, with \ourmetric, we can separately investigate different aspects.
While the classification performance follows the same trend as the TAO metric, the localization and association scores are still stable.
This allows us to instantly understand that the degradation is due to classification.

\begin{figure}[!tb]
    
    \begin{minipage}{.56\textwidth}
        \includegraphics[width=.95\linewidth]{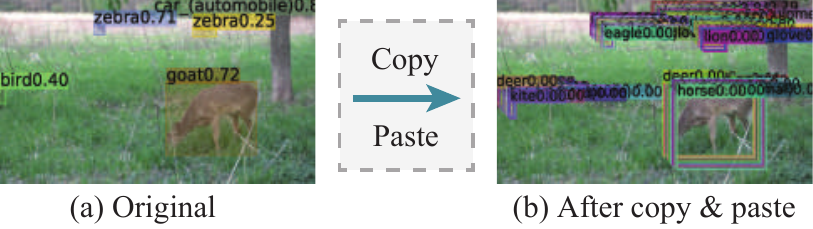}
    \end{minipage}%
    \begin{minipage}{0.4\textwidth}
    \scalebox{0.6}{
    \begin{tabular}{c|cccc}
	    \specialrule{.1em}{.05em}{.05em} 
    \rowcolor{mygray}TAO track mAP & AP$\uparrow$   & AP$_{50} \uparrow$   & AP$_{75} \uparrow$     & AR $\uparrow$    \\ \hline
    Before copy   & 0      & 0      & 0      & 0      \\
    After copy    & 62.9  & 75.2  & 50.5  & 62.9  \\ 
    \specialrule{.1em}{.05em}{.05em} 
    \rowcolor{mygray}Federated HOTA & HOTA $\uparrow$  & DetA $\uparrow$& AssA $\uparrow$&-   \\ \hline
    Before copy   & 4.2      & 3.0      & 5.9    &-  \\
    After copy    & 68.7  & 75.7  & 62.8  & - \\ 
    \specialrule{.1em}{.05em}{.05em} 
    \rowcolor{mygray}\ourmetric~(ours)           & TETA $\uparrow$   & LocA $\uparrow$  & AssocA $\uparrow$  & ClsA $\uparrow$  \\ \hline
    Before copy   & 47.6 & 80.1 & 59.6  & 3.2 \\
    After copy    & 13.8 & 3.3  & 10.2 & 27.9 \\

    \specialrule{.1em}{.05em}{.05em} 
    \end{tabular}
    }
    \end{minipage}
    \captionlistentry[table]{A table beside a figure}\label{tab:game_tao}
    \captionsetup{labelformat=andtable}
    \vspace{-0.3cm}
    \caption{Copy \& paste strategy to cheat the TAO track mAP. (a) tracking result from our tracker, which incorrectly classifies the deer as a goat. (b) copying and pasting existing tracks with low confidence class predictions from the object detector. The table shows the comparison between TAO track mAP and \ourmetric with a simple copy \& paste trick for the sequence in~\figurename~\ref{fig:game_tao}}
    \vspace{-0.4cm}
   \label{fig:game_tao}
\end{figure}

\parsection{Cheating TAO track mAP metric.}
\label{sec:trick_tao}
\figurename~\ref{fig:game_tao} shows a copy \& paste trick that can boost the TAO track mAP.
We simply copy and paste existing trajectories with low confidence class predictions from the object detector without additional training.
As shown in~\tablename~\ref{tab:game_tao}, TAO track mAP and Federated HOTA metric increase drastically from 0 to 62.9 and 4.2 to 68.7. 
In comparison, TETA drops from 47.6 to 13.8, which suggests the trick does not work on TETA.
Moreover, we can clearly see consequences brought by copy \& paste.
The localization score drops sharply as copy \& paste generates a lot of false positive localizations.
On the other hand, the trick only improves the classification performance.

\begin{table}[!tb]

\begin{minipage}[t]{.4\textwidth}
  \centering
  \caption{Results on TAO}  
 
  \scalebox{0.65}{
  \begin{tabular}{l|cccc}
    \specialrule{.1em}{.05em}{.05em} 
    Method & TETA & LocA & AssocA & ClsA \\
    \hline
    SORT~\cite{bewley2016simple} & 24.845 & 48.13 & 14.32 & 12.08 \\
    Tracktor~\cite{bergmann2019tracking} & 24.15  & 47.41 & 12.96 & 12.08 \\
    DeepSORT~\cite{wojke2017simple} & 25.98  & 48.35 & 17.52 & 12.09 \\
    AOA~\cite{du20211st} & 25.27  & 23.40 & 30.56 & \textbf{21.86} \\
    Tracktor++~\cite{bergmann2019tracking} & 27.97  & 49.04 & 22.81 & 12.05 \\
    QDTrack~\cite{qdtrack} & 30.00  & 50.53 & 27.36 & 12.11 \\
    \rowcolor{mygray}\textbf{TETer} & \textbf{33.25} & \textbf{51.58} & \textbf{35.02} & 13.16 \\
    \hline
    QDTrack-SwinT & 31.22  & 51.32 & 27.27 & 15.06 \\
    \rowcolor{mygray}\textbf{TETer-SwinT} & \textbf{34.61} & \textbf{52.10} & \textbf{36.71} & 15.03 \\
    \hline
    QDTrack-HTC & 32.79  & 56.21 & 27.43 & 14.73 \\
    \rowcolor{mygray}\textbf{TETer-HTC} & \textbf{36.85} & \textbf{57.53} & \textbf{37.53} & 15.70 \\
    \specialrule{.1em}{.05em}{.05em} 
  \end{tabular}}
  \label{tab:tao_res}
    \end{minipage}%
\begin{minipage}[t]{0.6\textwidth}
    \centering
    \caption{Results on BDD100K} 
    \label{tab:bdd_res}
    \scalebox{0.7}
    {\begin{tabular}{l|c|cc|cccc}
    \specialrule{.1em}{.05em}{.05em} 
    Method     & Split & mMOTA & mIDF1 & TETA & LocA & AssocA & ClsA\\  
    \hline
    DeepSORT~\cite{wojke2017simple} & val & 35.2 & 49.3 & 48.03 & 46.36 & 46.69 & 51.04 \\ 

    QDTrack~\cite{qdtrack}    & val   & 36.6          & 51.6  & 47.84 & 45.86 & 48.47 & 49.20 \\
    \rowcolor{mygray}\textbf{TETer}       & val   & \textbf{39.1} & \textbf{53.3} & \textbf{50.83} & \textbf{47.16} & \textbf{52.89} & \textbf{52.44} \\ 
    \hline
    DeepSORT~\cite{wojke2017simple} & test & 34.0 & 50.2 & 46.75 & 45.26 & 47.04 & 47.93 \\
    QDTrack~\cite{qdtrack}    & test  & 35.7 & 52.3  & 49.17 & \textbf{47.19} & 50.93 & 49.38 \\
    \rowcolor{mygray}\textbf{TETer }  & test  & \textbf{37.4} & \textbf{53.3} & \textbf{50.42} & 46.99 & \textbf{53.56} & \textbf{50.71} \\
    \specialrule{.1em}{.05em}{.05em} 
    \end{tabular}
    }
    \end{minipage}

\end{table}

\subsection{TAO Tracking Results}

We provide a thorough comparison of TETer against competing methods on the TAO validation set in~\tableautorefname~\ref{tab:tao_res} using our \ourmetric metric. We set the margin $r$ of local clusters to 0.5 since we observe the TAO dataset is not crowded, and this choice gives a proper balance between non-penalizing and over-penalizing FPs.
For this experiment, we only use the predefined 302 categories without considering the unknown classes. We allow each tracker maximum outputs 50 predictions per image. We use the same FasterRCNN detectors with class-agnostic NMS for all methods except AOA~\cite{aoa}.
Despite the increased difficulty introduced by the large number of categories, TETer outperforms all other methods, providing consistent improvements in TETA, LocA, and AssocA. In particular, TETer improves TETA of QDTrack~\cite{qdtrack} by over 3 points and AssocA by over 7 points. \par
We also compare our method to AOA~\cite{du20211st}, the winner of the ECCV 2020 TAO challenge, using the publicly available predictions\footnote{\url{https://github.com/feiaxyt/Winner\_ECCV20\_TAO}}. 
AOA combines multiple state-of-the-art few-shot detection and object ReID models that are trained using additional external datasets, which enables it to obtain very strong classification performance.
However, as it is optimized using the TAO metric, it makes excessive false positive predictions, which are punished by \ourmetric.
Additionally, TETer achieves better association performance without using external datasets.

\subsection{BDD100K Tracking Results}

We provide evaluation results on both the BDD100K validation and test sets in~\tableautorefname~\ref{tab:bdd_res}.
We first evaluate each tracker using the established CLEAR MOT metrics, including MOTA, MOTP, and IDF1, each averaged over every class.
Without bells and whistles, TETer can obtain performance gains across all three metrics compared to QDTrack~\cite{qdtrack} on both sets, achieving state-of-the-art performances.
In particular, TETer improves the mMOTA and mIDF1 of QDTrack by 2.5 and 1.7 points on the validation set and 1.7 and 1 points on the test set.
We also show evaluation results using \ourmetric.
TETer again obtains consistent performance gains across all metrics.
On both validation and test sets, TETer can improve AssocA of QDTrack by over 2.5 points.

\begin{figure}[!tb]

    \begin{minipage}{.5\textwidth}
        \centering
        \includegraphics[width=\linewidth]{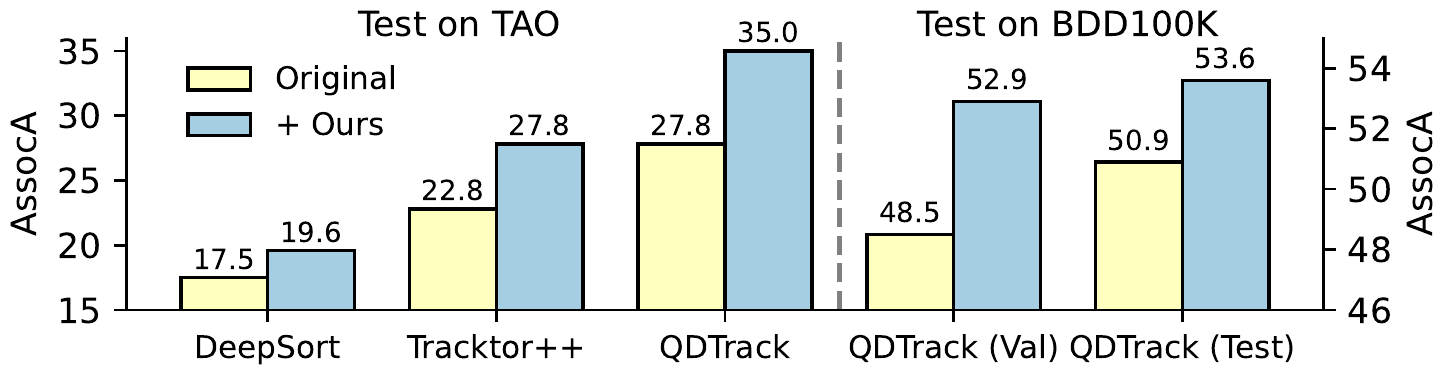}
    \end{minipage}
    \begin{minipage}{0.5\textwidth}
      \scalebox{0.6}{
  \begin{tabular}{@{\extracolsep{\fill}}l|cccc|ccc}
    \specialrule{.1em}{.05em}{.05em} 
    & \multicolumn{4}{c|}{Known} & \multicolumn{3}{c}{Unknown} \\ 
    Components  & TETA & LocA & AssocA & ClsA & TETA & LocA & AssocA  \\
    \hline
    QDTrack~\cite{qdtrack} & 28.89 & 46.60 & 27.76 & 12.37 & 19.89 & 32.25 & 27.42 \\
    + CAL  & 30.00 & 50.53 & 27.36 & 12.11 & 20.85 & 36.06 & 26.50 \\
    + CEM  & 32.86 & 51.55 & 35.00 & 12.06 & 24.00 & 37.38 & 34.61 \\
    + TCC  & \textbf{33.25} & \textbf{51.55} & \textbf{35.00} & \textbf{13.16} & \textbf{24.00} & \textbf{37.38} & \textbf{34.61} \\
    \specialrule{.1em}{.05em}{.05em}
  \end{tabular}
  }
  \end{minipage}\hspace{0.3cm}
    \captionlistentry[table]{A table beside a figure}\label{tab:open-set}
    \captionsetup{labelformat=andtable}
    \caption{~\figurename~\ref{fig:cem_improve} shows comparison of DeepSORT~\cite{he2016deep}, Tracktor++~\cite{bergmann2019tracking}, and QDTrack~\cite{qdtrack} with w/ and w/o our CEM module on TAO and BDD100K datasets. CEM consistently improves association performance of all methods. ~\tablename~\ref{tab:open-set} shows comparison with different components of TETer on the TAO open set using our \ourmetric metrics}
   \label{fig:cem_improve}
\end{figure}

\subsection{Generalizability of CEM}

To demonstrate the generalizability of our CEM module,
we further apply CEM to other MOT methods to replace existing hard prior-based matching.
We compare three methods, DeepSORT~\cite{wojke2017simple}, Tracktor++~\cite{bergmann2019tracking}, and QDTrack~\cite{qdtrack}, with and without CEM across both TAO and BDD100K.
The results are shown in~\figurename~\ref{fig:cem_improve}.
On the TAO validation set, adding CEM results in at least 2 points of improvement in AssocA across all methods.
In particular, CEM can improve AssocA of QDTrack by 7 points.
On both BDD100K validation and test sets, CEM can obtain over 2.5 points of improvement in AssocA of QDTrack.
This shows our CEM module can be applied to various popular MOT methods and achieve consistent improvements in association by better exploiting class information.
\subsection{Ablation Study}
We conduct ablation studies on TAO and BDD100K.
We investigate the importance of our proposed modules on both predefined and unknown categories. For TAO, we use their split for known and unknown (free-form) classes. For unknown split, we only report the LocA and AssocA.

\parsection{Tracking Components.}
We evaluate the contributions of each component of TETer on the TAO open set using \ourmetric in~\tablename~\ref{tab:open-set}. When we replace the class-dependent intra-class NMS from the object detector with a class-agnostic NMS, we can improve LocA by over 3 points on both known objects and unknown objects.
Adding CEM drastically improves its AssocA by over 7 points on known objects and 8 points on unknown objects.
Further, using temporal class correction can improve ClsA by over 1 point.

\parsection{Comparison of using class information.}
\label{sec:aet_baseline}
We compare different ways of utilizing class information during association on the validation set of BDD100K and TAO in~\tablename~\ref{tab:paradigm_compare}.
The baseline protocol follows the AET strategy and performs class-agnostic association with pure instance appearance features described in~\sectionautorefname~\ref{sec:asso}. We then add different class prior on top of the AET baseline to study their effectiveness.
Softmax use class labels as hard prior and associate objects within the same class. This strategy leads to a severe downgrade in the tracking performance, especially for the TAO dataset.

Alternatively, we use the out-of-the-shelf word embeddings to incorporate class information. The semantically similar classes should be closer in the word embedding space. This way transfers the hard class labels to soft ones. We utilize the BERT model to embed the class names to replace our CEM. While the performance is slightly better than using softmax predictions, it is inferior to the CEM. Our CEM is the only method capable of effectively utilizing semantic information to improve the association by outperforming the AET baseline on large-scale long-tailed datasets.

\begin{table}[tb]
\centering
\caption{
Comparing different ways of using class information. AET Baseline associates every objects without using any class information. Softmax indicates association happens only within the same class. BERT indicates using BERT word embeddings to group candidates for association}
\footnotesize
\scalebox{0.85}{
\begin{tabular*}{\textwidth}{@{\extracolsep{\fill}}l|ccc|c}

\specialrule{.1em}{.05em}{.05em} 
         & \multicolumn{3}{c|}{BDD100K val}   & \multicolumn{1}{c}{TAO val} \\
Class & mMOTA & mIDF1 & AssocA  & AssocA                \\ 
\hline
AET Baseline
& 37.3  & 52.6  & 52.0   & 33.5 \\
Softmax &  36.6~(\textcolor{red}{-0.7})   & 51.6~(\textcolor{red}{-0.2})  & 48.9~(\textcolor{red}{-0.2})  &  27.4~(\textcolor{red}{-6.1})       \\
BERT & 37.2~(\textcolor{red}{-0.1})  & 52.6   & 52.0 &  27.8~(\textcolor{red}{-5.7})              \\
CEM & \textbf{39.1}~(\textcolor{cyan}{+1.8})   & \textbf{53.3}~(\textcolor{cyan}{+0.7})   & \textbf{52.9} ~(\textcolor{cyan}{+0.9})   & \textbf{35.0} ~(\textcolor{cyan}{+1.5})         \\
\specialrule{.1em}{.05em}{.05em}
\end{tabular*}}
\label{tab:paradigm_compare}

\end{table}

\section{Conclusion}

We present a new metric TETA and a new model TETer for tracking every thing in the wild. TETA and TETer disentangle classification from evaluation and model design for the long-tailed MOT. TETA can evaluate trackers more comprehensively and better deal with incomplete annotation issues in large-scale tracking datasets. TETer disentangles the unreliable classifier from both detection and association, resulting in a better tracker which outperforms existing state-of-the-art trackers on large-scale MOT datasets, TAO and BDD100K. The core component of TETer, CEM, can be used as a drop-in module for existing tracking methods and boost their performance.

\section{Acknowledgement}
Special thanks go to Ruolan Xiang for her help in editing the paper.

%% file: appendix.tex
\section{Appendix}
We give more details on the ablation study regarding our Tracking-Every-Thing Accuracy (TETA) metric and our Tracking-Every-Thing tracker (TETer) and provides additional evaluation training and implementation details. 

\subsection{BDD100K Per-class Evaluation Results}

We provide per-class evaluation results using CLEARMOT~\cite{MOTA} and TETA metrics on the BDD100K~\cite{bdd100k} validation set in~\tableautorefname~\ref{tab:bdd-perclass-val}.
Data distribution in BDD100k is long-tailed. The Car category consists of most of the tracks in the dataset. The rest of the categories are rare compared to the dominant ones. Thus, we characterized them as rare classes.
TETer can achieve significant improvements across all rare classes on both established MOTA, IDF1, and our TETA metrics compared to the previous state-of-the-art QDtrack.
In particular, TETer boosts MOTA of buses by over 7 points on the validation set and TETA by over 6 points. 
We also compare our TETer results with CEM with its class agnostic counterpart, where the model only uses the AET strategy without CEM. The result shows that our model gains significant improvements over rare classes where the class agnostic instance association cannot be well trained due to lacking annotations. For instance, we gain +3.8 MOTA on buses and +4.7 MOTA on riders. Further, we can observe improvements on the TETA score, where we gain +2 on train and +1.5 on motorcycle. This demonstrates that TETer can better handle tracking rare classes. With CEM, we exploit the semantic annotations offered by large-scale object detection datasets. It can integrate fine-grained cues required for classification (\eg the difference between a big red bus and a red truck), which are difficult to learn effectively with class-agnostic appearance training on the long-tailed datasets.

\begin{table*}[t]
\centering
\caption{Per-class evaluation results on the BDD100K validation set using CLEARMOT and TETA metrics. Rare classes are highlighted in gray}
\label{tab:bdd-perclass-val}

\begin{tabular}{@{}l|l|ccc|cccc@{}}
\toprule
Method & Category & MOTA & MOTP & IDF1 & TETA & LocA & AssocA & ClsA \\ \hline
\multirow{9}{*}{QDTrack~\cite{qdtrack}} 
& Pedestrian & 49.3 & 78.4 & 59.9 & 52.1 &50.9 &46.8 &58.7 \\
 & Rider & 35.0 & 77.5 & 51.5 &45.1 &47.3 &39.6 &48.5 \\
 & Car & 69.8 & 84.6 & 75.0 &69.1 &62.2 &65.6 &79.5 \\
 & Truck & 39.2 & 85.4 & 58.2 &55.5 &57.0 &55.6 &53.9 \\
 & Bus & 40.8 & 86.2 & 62.3 &57.9 &58.1 &57.5 &58.1 \\
 & Train & 0.0 & - & 0.0 &12.1 &15.6 &20.7 &0.0 \\
 & Motorcycle & 28.8 & 76.9 & 56.0 &46.4 &41.6 &54.0 &43.5 \\
 & Bicycle & 30.0 & 76.2 & 50.1  &44.6 &34.2 &48.0 &51.4 \\\cline{2-9} 
 & Average & 36.6 & 70.7 & 51.6 &47.8 &45.9 &48.5 &49.2 \\ \hline
  \multirow{9}{*}{AET (Class-agnostic)}
&Pedestrian &47.8 &79.1 &59.3 &53.0 &52.7 &47.0 &59.4 \\
&Rider &35.9 &76.4 &53.2 &48.2 &51.6 &48.1 &45.0 \\
&Car &69.6 &85.4 &75.0 &70.5 &64.1 &66.5 &80.8 \\
&Truck &41.7 &85.2 &59.8 &59.5 &59.0 &63.2 &56.1 \\
&Bus &44.4 &86.1 &66.2 &61.8 &61.1 &65.4 &58.8 \\
&Train &-2.6 &- &0.0 &12.4 &15.7 &21.4 &0.0 \\
&Motorcycle &31.6 &76.8 &57.8 &47.0 &42.8 &55.3 &42.8 \\
&Bicycle &29.8 &76.9 &49.7 &45.5 &35.2 &48.7 &52.4 \\\cline{2-9} 
 & Average   & 37.3 & 70.7 & 52.6 &49.7 &47.8 &52.0 &49.4 \\ \hline
 \multirow{9}{*}{\textbf{TETer~(CEM)}} 
&Pedestrian &49.7 &79.1 &59.9 &54.1 &52.3 &47.9 &62.2 \\
&Rider &40.5 &76.5 &56.6 &49.9 &50.4 &49.3 &50.1 \\
&Car &69.7 &85.4 &74.2 &70.5 &63.8 &65.6 &82.1 \\
&Truck &43.3 &85.5 &59.7 &59.1 &58.2 &60.1 &59.2 \\
&Bus &48.2 &86.2 &67.6 &63.7 &60.0 &65.8 &65.3 \\
&Train &0.0 &- &0.0 &14.4 &15.2 &28.1 &0.0 \\
&Motorcycle &31.6 &77.0 &58.2 &48.5 &42.8 &56.3 &46.6 \\
&Bicycle &30.1 &77.1 &50.1 &46.3 &34.7 &50.0 &54.2 \\
\cline{2-9} 
 & Average & \textbf{39.1} & \textbf{70.8} & \textbf{53.3} & \textbf{50.8} & \textbf{47.2} & \textbf{52.9} & \textbf{52.4} \\ \bottomrule
\end{tabular} 

\end{table*}

\begin{table}[t]
\centering
\caption{Changing the margin $r$ with a fixed $\alpha$}
\begin{tabular}{@{}l|c|cccccc@{}}

\hline
Method                         & $r$  & TETA   & LocA   & AssocA & ClsA & LocRe & LocPr    \\ \hline
\multirow{3}{*}{QDTrack}       & 50 & 30.0 & 50.5 & 27.4 & 12.1 & 53.1 & 75.8 \\
                               & 75 & 30.6 & 52.4 & 27.4 & 12.1 & 53.1 & 84.7 \\
                              & 90 & 30.8 & 53.0 & 27.4 & 12.1 & 53.1 & 91.1 \\ \hline
\multirow{3}{*}{\textbf{TETer (ours)}}  & 50 & 33.2 & 51.6 & 35.0 & 13.2 & 54.3 & 75.3 \\
                              & 75 & 33.9 & 53.6 & 35.0 & 13.2 & 54.3 & 84.3\\
                              & 90 & 34.1 & 54.1 & 35.0 & 13.2 & 54.3 & 90.5\\ \hline
\end{tabular}
\label{tab:fixa}
\end{table}

\begin{figure}[t]
	\centering
	\includegraphics[width=0.5\linewidth]{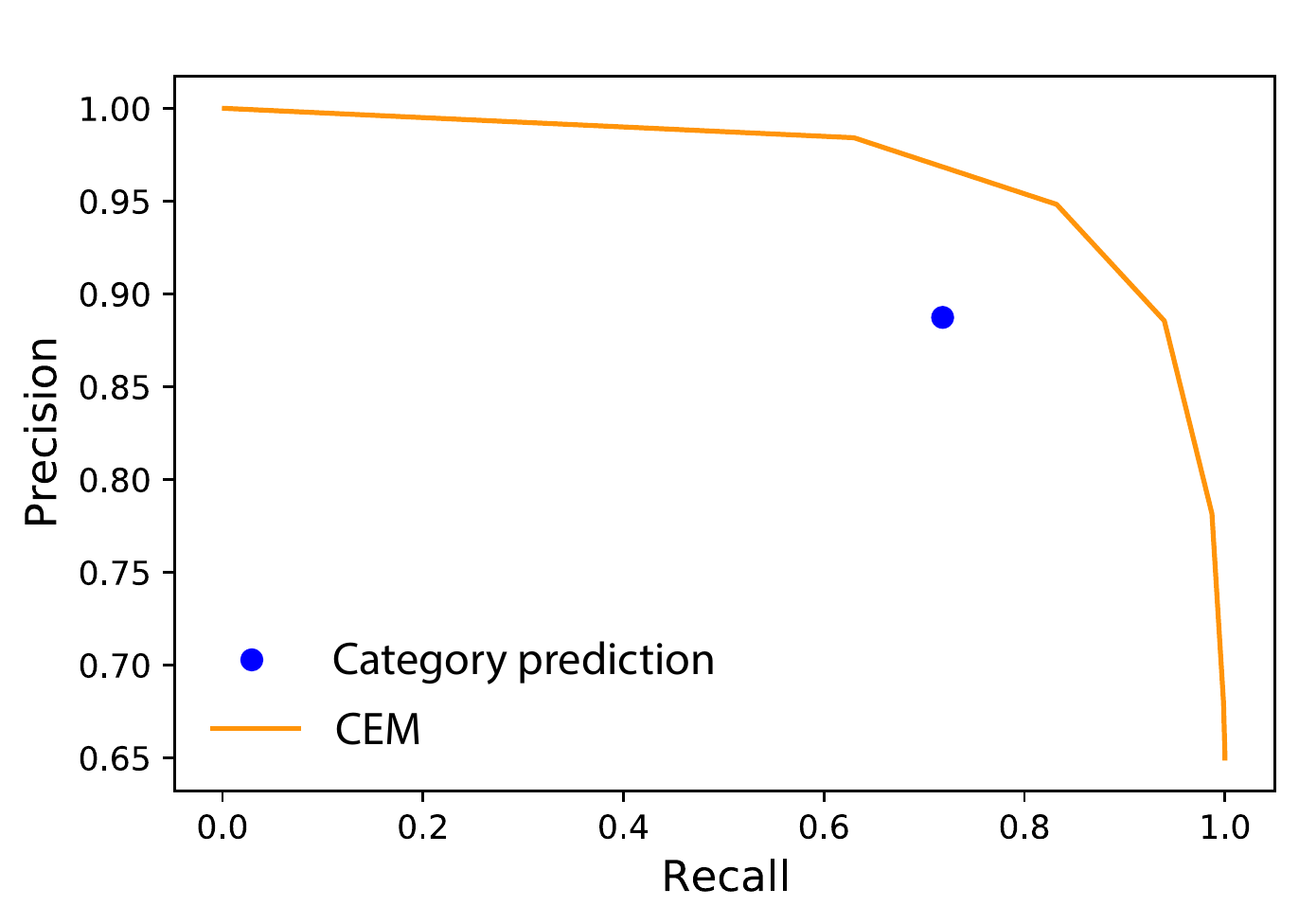}
	\caption{Precision-Recall (PR) curves of in-sequence instance retrieval based on CEM and category prediction on the TAO validation set. A retrieved result is correct if it has the same labeled category with the target instance }
	\label{fig:PR_TAO_FULL}
\end{figure}

\subsection{TAO Per Frequency Group Evaluation Results}

We provide evaluation results per frequency group on the TAO validation set using TETA in~\tableautorefname~\ref{tab:tao-freq}.
We first observe that TETA can effectively evaluate methods across different frequency groups, despite difficulties introduced by classification errors.
Although ClsA drops significantly for both QDTrack~\cite{qdtrack} and TETer as categories become more rare, LocA and AssocA are relatively stable.
This enables us to compare different methods even in large-scale, long-tailed settings where classification is the bottleneck.

Compared to QDTrack, TETer can obtain consistent improvements in TETA, LocA, and AssocA across all frequency groups, at the cost of a small degradation in ClsA.
The improvements are more prominent on common and rare categories, where TETer can achieve over 3 points improvement in TETA.
For rare categories, TETer achieves 1.8 points improvement in LocA and over 7 points in AssocA.
Even on frequency categories, TETer can still improve AssocA by over 6 points.
~\tableautorefname~\ref{tab:tao-freq} also shows that the major differences between frequent and rare categories lies in classification. The localization and association capabilities of both trackers already generalize very well on rare categories.

\begin{table}[t]
\centering
\caption{Per frequency group results on the TAO validation set using TETA}
\label{tab:tao-freq}
\small
\begin{tabular}{@{}l|l|cccc@{}}
\toprule
Method & Freq. Group & TETA & LocA & AssocA & ClsA \\ \hline
\multirow{4}{*}{QDTrack~\cite{qdtrack}}
& Frequent & 36.3 & 52.4 & 32.0 & 24.5 \\
 & Common & 23.9 & 47.2 & 21.7 & 2.9 \\
 & Rare & 26.7 & 52.7 & 27.4 & 0.0 \\
 & All & 30.0 & 50.5 & 27.4 & 12.1 \\ \hline
\multirow{4}{*}{\textbf{TETer~(ours)}}
 & Frequent & 39.4 & 53.9 & 38.7 & 25.7 \\
 & Common & 27.3 & 47.3 & 30.4 & 4.1 \\
 & Rare & 30.1 & 54.5 & 35.3 & 0.4 \\
 & All & \textbf{33.2} & \textbf{51.6} & \textbf{35.0} & \textbf{13.2} \\ \bottomrule
\end{tabular}
\end{table}


\subsection{Exemplar-based Classification}

Given an example object, exemplar-based classification means classifying objects by comparing with the given example to determine whether they belong to the same class.
Given two neighboring frames $t_1$ and $t_2$ in a video sequence, all objects in $t_1$ will be treated as exemplars.
For each exemplar, we find all target objects in $t_2$ that belong to the same class as the exemplar.

In this experiment, we compare our Class Exemplar Matching (CEM) with a hard prior baseline that matches objects with the same predicted class label.
We evaluate both methods on the TAO validation set
and compute precision-recall (PR) curves for comparison.
A true positive (TP) match is a match between two objects that belong to the same category.
A false positive (FP) match is a match between two objects that belong to different categories.
A false negative (FN) is a non-match between two objects that belong to the same category.
To compute the PR curve, we sample 10 thresholds from 0 to 0.99 with a fixed step size.

\figurename~\ref{fig:PR_TAO_FULL} shows the results of the experiment.
The hard prior baseline takes the argmax of the predictions of a softmax classifier from Faster R-CNN, thus there is only a single value in the PR curve. CEM significantly outperforms the hard prior baseline.

\subsection{TETA Details}

We provide additional details regarding our TETA metric about how it disentangles classification and how it deals with evaluation on datasets with complete annotations.

\subsubsection{Disentangling Classification}

The most direct way to disentangle classification is not to consider per-class performance and evaluate every object class-agnostically.
However, on large-scale, long-tailed datasets, such evaluation will be dominated by objects of the few common categories,
and the overall performance will not reflect the improvements on rare classes.
On the other hand, per-class evaluation requires us to select prediction results for each class, which is sensitive to classification performance.
If the classification is wrong, the contribution in localization and association will be ignored. 
TETA can naturally deal with this issue with the local cluster evaluation since we select predictions based on their location rather than class.
To evaluate a particular class, we access predictions in the local clusters of ground-truth objects belonging to the chosen class.
Thus, we can evaluate the localization and association performance even when the class predictions are wrong.

\parsection{Complete annotations: multiple categories}
TETA can also work with complete annotations. First, the localization accuracy is not affected.
In the case of incomplete annotations, we treat every unmatched predictions in each cluster as false positives.
If we have exhaustive annotations, we still treat those unmatched predictions as false positives.
The remaining question is how to penalize predictions that are not in any clusters.
For such predictions,
we know that they are not highly overlapped with any ground truth box, since we have exhaustive annotations.
The predictions thus false classify background as one of the foreground classes,
and so we treat them as classification false positives.

\parsection{Complete annotations: single category}
For single category with exhaustive annotations, the classification term of TETA is meaningless and can be ignored.
Also, since we do not need to perform per-class evaluation, the margin of the local cluster does not matter either.
Thus, we can set the margin $r$ to 0.
With these changes, TETA becomes similar to the HOTA~\cite{hota} metric with the only difference being that we use arithmetic mean instead of geometric mean.

\parsection{Ablation Study of TETA}
We provide an ablation study of the local cluster IoU margin $r$ of TETA. We perform this experiment on incomplete dataset TAO.
The results are shown in~\tablename~\ref{tab:fixa}. The LocRe and LocPr represent the localization recall and precision. As we can see, with a larger $r$, the LocPr increases since TETA becomes more conservative regarding identifying FPs. In the mean time, TETA makes fewer mistakes where the objects with no annotations are wrongly identified as FPs. In extreme crowded scenarios with incomplete annotations, it's recommended to set a higher $r$ to avoid false punishment.

\subsection{Qualitative Results}

We provide additional qualitative results of TETer.

\begin{figure*}[t]
	\centering
	\includegraphics[width=0.9\linewidth]{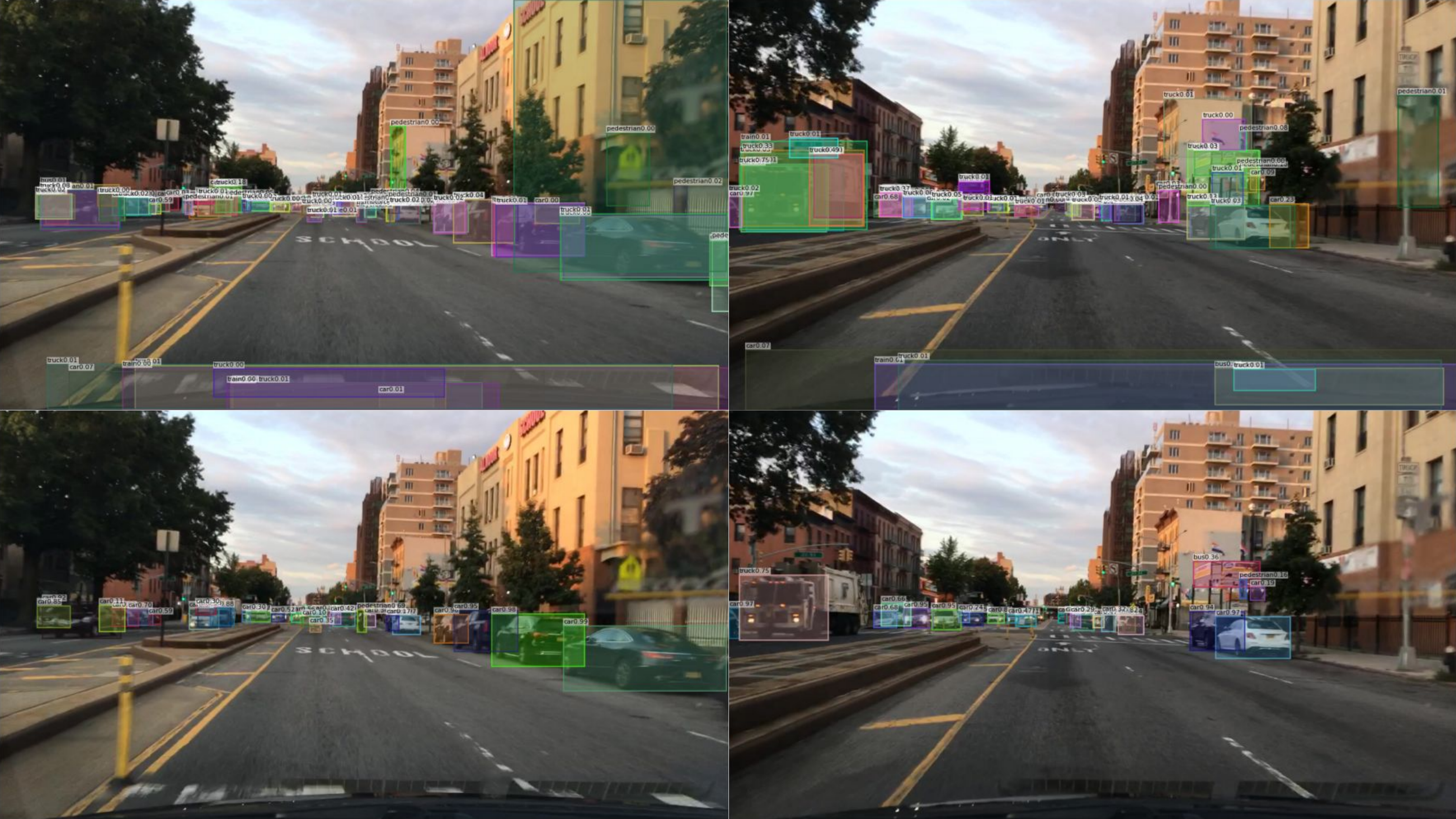}
	\vspace{-0.12in}
	\caption{Qualitative comparison of tracker optimized for the TAO~\cite{tao} metric (top) and tracker optimized for TETA (bottom) on BDD100K. The tracker optimized for TAO produces more false positives}
	\label{fig:sup-crossdataset}
\end{figure*}

\parsection{Category label prediction vs. CEM}
\begin{figure*}[t]
	\centering
	\includegraphics[width=0.9\linewidth]{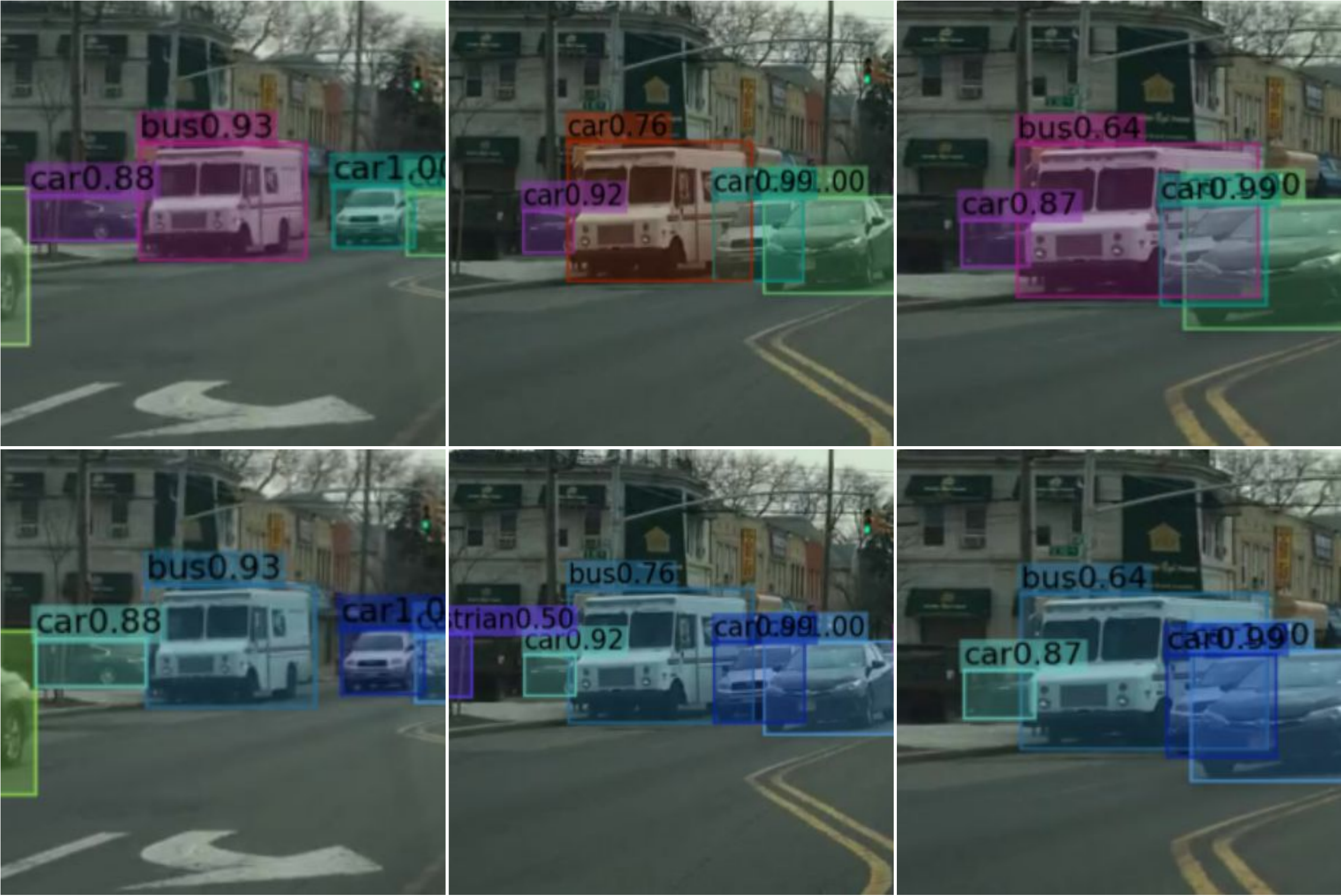}
	\vspace{-0.12in}
 	\caption{Qualitative comparison of QDTrack~\cite{qdtrack} (top) and TETer (bottom) on BDD100K. QDTrack has ID Switches due to classification errors. (Same color represent the same track)}
	\label{fig:sup-idswitch}
\end{figure*}
We first compare the QDTrack~\cite{qdtrack} which uses class prediction as hard prior to associate objects with TETer which uses CEM.
In~\figurename~\ref{fig:sup-idswitch}, we show an example of QDTrack producing ID Switches due to errors in classification, whereas TETer is more robust to such issues.

\begin{figure*}[t]
\centering
	\footnotesize
	\setlength{\tabcolsep}{0.1em}
	\adjustbox{width=1\linewidth}{
		\begin{tabular}{cccc}
		   \rotatebox{90}{\hspace{7mm}Class-agnostic}&\includegraphics[width=0.3\linewidth]{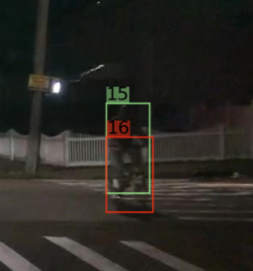}& \includegraphics[width=0.3\linewidth]{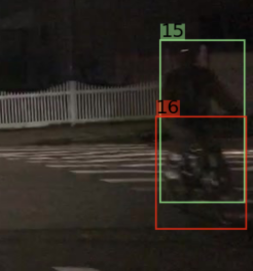}& \includegraphics[width=0.3\linewidth]{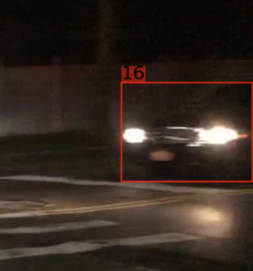} \\
   		   \rotatebox{90}{\hspace{14mm}CEM}&\includegraphics[width=0.3\linewidth]{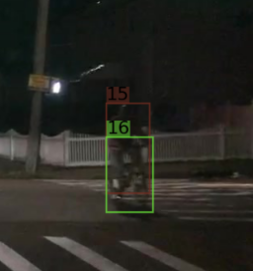}& \includegraphics[width=0.3\linewidth]{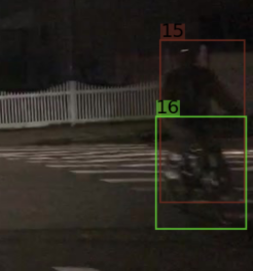}& \includegraphics[width=0.3\linewidth]{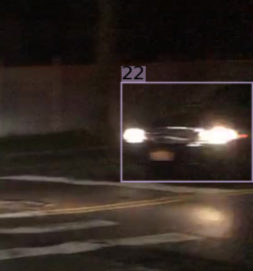} \\
   		   \multicolumn{4}{c}{(a) The class agnostic tracker false associates a bicycle (16) to a car (16). }\\
		   \rotatebox{90}{\hspace{7mm}Class-agnostic}&\includegraphics[width=0.3\linewidth]{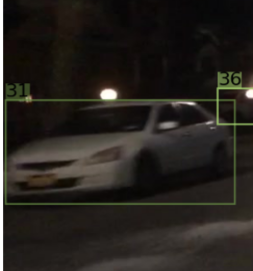}& \includegraphics[width=0.3\linewidth]{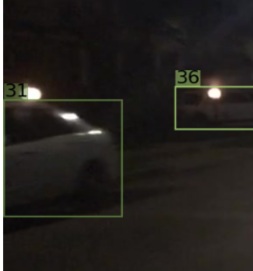}& \includegraphics[width=0.3\linewidth]{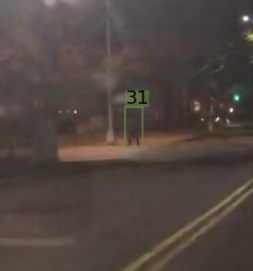} \\
   		   \rotatebox{90}{\hspace{14mm}CEM}&\includegraphics[width=0.3\linewidth]{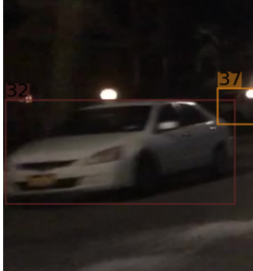}& \includegraphics[width=0.3\linewidth]{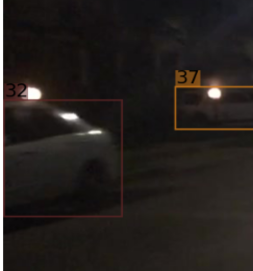}& \includegraphics[width=0.3\linewidth]{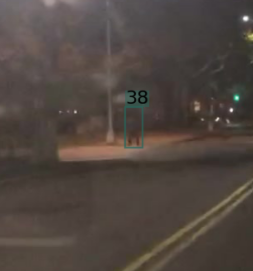} \\	
   		      		   \multicolumn{4}{c}{(b) The class agnostic tracker false associates a car (31) to a pedestrian (31). }\\
   		      		   
	\end{tabular}}

    \caption{The qualitative comparison between CEM and class-agnostic association. CEM can exploit the semantic supervision offered by large scale datasets to learn fine-grained class appearance differences. Therefore, it can help trackers to avoid the association mistakes in class-agnostic association}
    \label{fig:clsag_cem}
\end{figure*}

\parsection{Class-agnostic vs. CEM}
We further show the comparison between the class-agnostic association (AET baseline in Section 5.5) and association with CEM. We observe that most class agnostic association errors happen in rare classes where there are not enough videos to train the class-agnostic instance association module well. For instance, ~\figurename~\ref{fig:clsag_cem} (a) shows the bicycle (16) is wrongly associated with the car with class-agnostic association, while using the CEM module helps to avoid the mistake.
The CEM module utilizes the supervision from large-scale object detection datasets to learn fine-grained class appearance differences, which helps the association on rare classes.

\begin{figure*}[ht]
	\centering
	\includegraphics[width=1\linewidth]{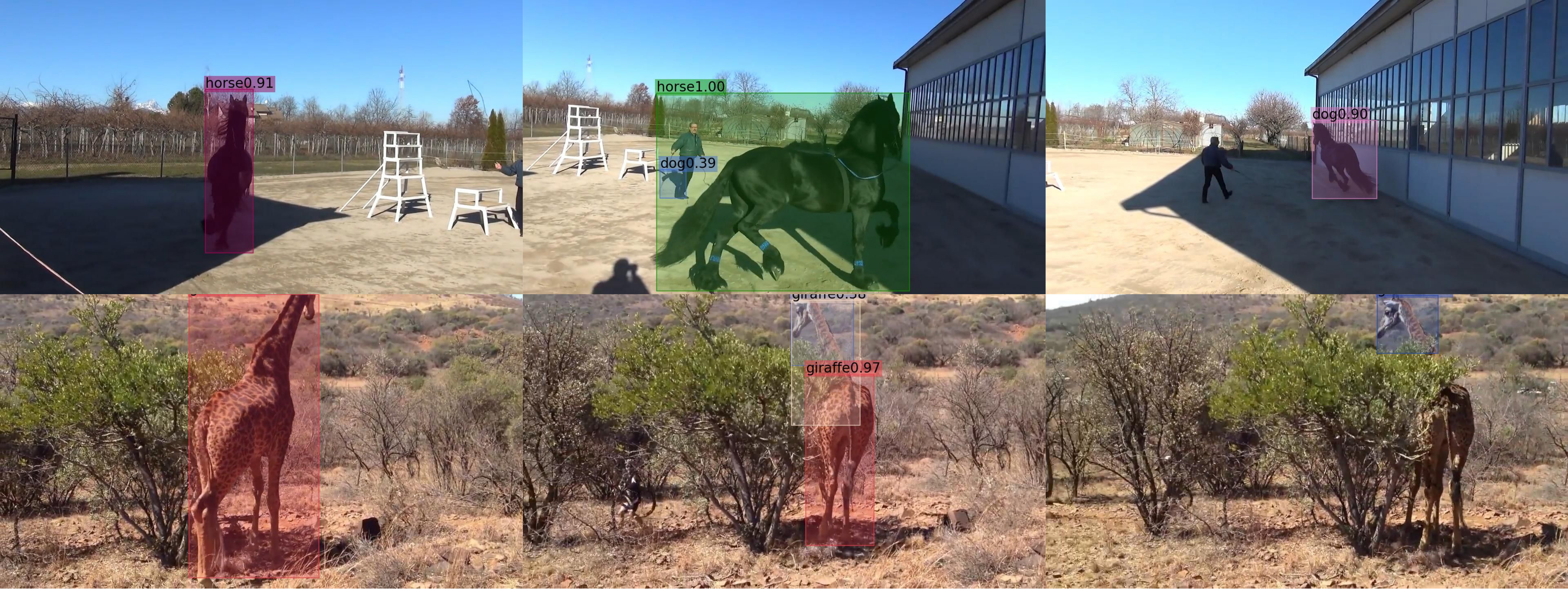}
	\vspace{-0.12in}
	\caption{Failure cases of TETer on TAO. Big appearance changes (top) and occlusions (bottom)}
	\label{fig:sup-failure}
\end{figure*}

\parsection{Tracking results comparison: TAO metric vs. TETA }
We provide results for cross-dataset analysis.
In~\figurename~\ref{fig:sup-crossdataset}, we show predictions from trackers that are optimized either for the TAO metric or TETA. The tracker optimized for the TAO metric generates more false positives that highly overlap, producing results that are difficult to use in practice. On the other hand, the tracker optimized for TETA produces cleaner results.

\begin{figure*}[t]
\centering
	\footnotesize
	\setlength{\tabcolsep}{0.1em}
	\adjustbox{width=1\linewidth}{
		\begin{tabular}{ccccc}
		   \rotatebox{90}{\hspace{10mm}Softmax}&\includegraphics[width=0.29\linewidth]{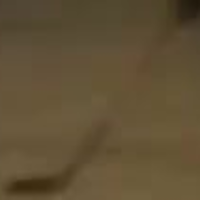}& \includegraphics[width=0.29\linewidth]{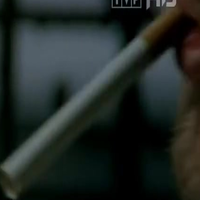}& \includegraphics[width=0.29\linewidth]{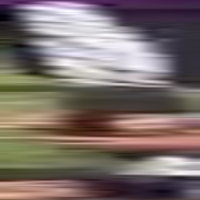}& \includegraphics[width=0.29\linewidth]{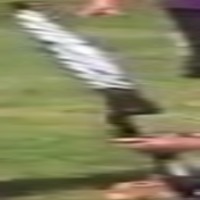} \\
		   &hockey stick (target)&cigarette&bow (weapon)&bow (weapon)\\
   		   \rotatebox{90}{\hspace{14mm}CEM}&\includegraphics[width=0.29\linewidth]{figs/supp/clsag_cem/hockey_stick-target.png}& \includegraphics[width=0.29\linewidth]{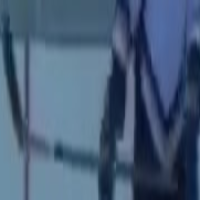}& \includegraphics[width=0.29\linewidth]{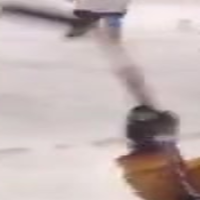}& \includegraphics[width=0.29\linewidth]{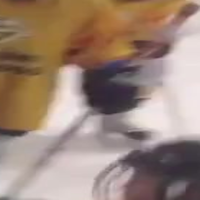} \\
   		   &hockey stick (target)&hockey stick&hockey stick&hockey stick\\
   		   \multicolumn{5}{c}{}\\
   		   \multicolumn{5}{c}{}\\
   		   \multicolumn{5}{c}{}\\
   		   \rotatebox{90}{\hspace{10mm}Softmax}&\includegraphics[width=0.29\linewidth]{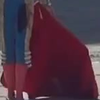}& \includegraphics[width=0.29\linewidth]{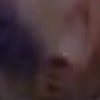}& \includegraphics[width=0.29\linewidth]{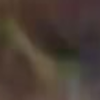}& \includegraphics[width=0.29\linewidth]{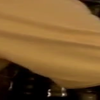} \\
   		   &cape (target)&grocery bag&ball&mixer\\

   		   \rotatebox{90}{\hspace{14mm}CEM}&\includegraphics[width=0.29\linewidth]{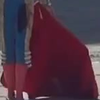}& \includegraphics[width=0.29\linewidth]{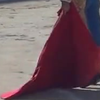}& \includegraphics[width=0.29\linewidth]{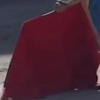}& \includegraphics[width=0.29\linewidth]{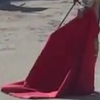} \\
		   &cape (target)&cape&cape&cape\\
	\end{tabular}}
	  \caption{Rare class retrieval. Objects in the most left column are the retrieval targets. The rest columns are the retrieval examples using softmax classifier or CEM. The retrieval results are ranked by confidence. Columns from left to right are corresponding to the results from high to low confidence. The class name under each example is the ground truth class for the example. The CEM module generalizes well on rare classes and can successfully retrieve them}
    
    \label{fig:cls_retrievl_good}
\end{figure*}

\begin{figure*}[t]
\centering
	\footnotesize
	\setlength{\tabcolsep}{0.1em}
	\adjustbox{width=1\linewidth}{
		\begin{tabular}{ccccc}
		   \rotatebox{90}{\hspace{10mm}Softmax}&\includegraphics[width=0.29\linewidth]{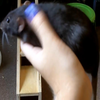}& \includegraphics[width=0.29\linewidth]{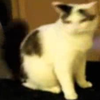}& \includegraphics[width=0.29\linewidth]{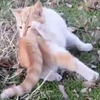}& \includegraphics[width=0.29\linewidth]{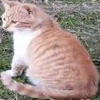} \\
		   &rat (target)&cat&cat&cat\\
   		   \rotatebox{90}{\hspace{14mm}CEM}&\includegraphics[width=0.29\linewidth]{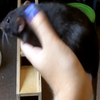}& \includegraphics[width=0.29\linewidth]{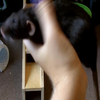}& \includegraphics[width=0.29\linewidth]{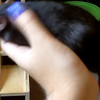}& \includegraphics[width=0.29\linewidth]{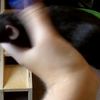} \\
   		   &rat (target)&rat&human&human\\
   		   
   		   \multicolumn{5}{c}{}\\
   		   \multicolumn{5}{c}{}\\
   		   \multicolumn{5}{c}{}\\
   		   \rotatebox{90}{\hspace{10mm}Softmax}&\includegraphics[width=0.29\linewidth]{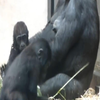}& \includegraphics[width=0.29\linewidth]{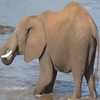}& \includegraphics[width=0.29\linewidth]{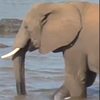}& \includegraphics[width=0.29\linewidth]{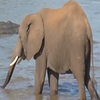} \\
   		   &gorilla (target)&elephant&elephant&elephant\\
   		   \rotatebox{90}{\hspace{14mm}CEM}&\includegraphics[width=0.29\linewidth]{figs/supp/clsag_cem/gorilla-target.png}& \includegraphics[width=0.29\linewidth]{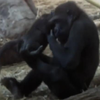}& \includegraphics[width=0.29\linewidth]{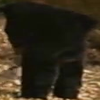}& \includegraphics[width=0.29\linewidth]{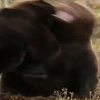} \\
		   &gorilla (target)&monkey&baby&bear\\

	\end{tabular}}

    \caption{Failure retrieval cases. The first two rows show that the CEM module retrieve the human hand as rat due to occlusion. The last two rows show the CEM retrieve monkey, bear as the gorilla due to the high visual similarities}
    
    \label{fig:cls_retrivel_bad}
\end{figure*}

\parsection{Rare class retrieval}
We perform the class retrieval experiments in the rare classes on TAO to show the effectiveness of the CEM embeddings. We take the objects in the first frame of each ground truth track and use them as the retrieval templates to retrieve ground truth objects in the whole TAO validation set. The softmax prediction means we use the softmax confidence to retrieve objects that are predicted as the same class as the template. The CEM means we use the CEM embedding similarity to perform the retrieval. ~\figurename~\ref{fig:cls_retrievl_good} shows the CEM embedding can successfully retrieve the examples in the rare classes, while its softmax fails. 
~\figurename~\ref{fig:cls_retrivel_bad} shows some failures cases where the CEM module retrieves the wrong class due to occlusion or high visual similarities.

\parsection{TETer failure cases}
We also show some common failure cases of TETer on TAO in~\figurename~\ref{fig:sup-failure}. Note that TAO is annotated at 1 FPS. Thus, fast-moving objects usually have huge appearance changes in neighboring frames.
Due to the large appearance and location variations, tracking is challenging on TAO. Also, TETer suffers from localization errors caused by occlusion.

\subsection{More Implementation Details}
\label{implement-details}
We provide more implementation and training details of our method and evaluation setup in different benchmarks.

\parsection{Network architecture}
We use the popular Faster R-CNN ~\cite{ren2015faster} with ResNet as the backbone. Specifically, we use the ResNet-101~\cite{he2016deep} for TETer on TAO and ResNet-50 on BDD100K. For the exemplar encoder, we use \textit{4conv-3fc} head with group normalization ~\cite{wu2018group}. The final output channel numbers are 1230 for TETer on TAO~\cite{tao} and 256 for BDD100K~\cite{bdd100k}. We use the same network architecture for the instance appearance encoder but with only \textit{1fc} layers for the final output. The channel number of the instance appearance encoder is 256 by default on both datasets.

\parsection{Train: TAO}
We train TETer following the TAO~\cite{dave2020tao} set up with a mixed LVISv0.5~\cite{lvis} and COCO~\cite{coco} dataset. We set the batch size to 16 and the learning rate to 0.02. We train 24 epochs in total and decrease the learning by 0.1 after 16 and 22 epochs. 
For data augmentation, we randomly flip the images horizontally with a 0.5 ratio. We randomly resize the training images to keep their short edges between 640 to 800. We randomly sample images to form mini-batches with additionally repeat sampling for rare classes~\cite{lvis}. We set the repeat factor to 0.001. We train the instance appearance encoder on the TAO training set following the same setting as QDTrack~\cite{qdtrack}.

\parsection{Train: BDD100K}
We use the same object detector as QDTrack~\cite{qdtrack}. For training the exemplar encoder, we freeze the object detector above and train with 8 BDD100K MOT categories using the BDD100K Detection set, which contains 70K images. For data augmentation, we randomly flip the images horizontally with a 0.5 ratio. We randomly resize the training images to keep their short edges between 640 to 800. We randomly sampled images to form mini-batches. We set the batch size to 128.

\parsection{Test: TAO}
We evaluate our model on the TAO validation set with TETA. For the close-set setting, the TAO validation set contains 988 videos with 302 classes, a subset with LVIS classes. For the open-set setting, we merge the additional free-form classes~\cite{tao} as one unknown class. During inference, we use the fixed image scale with 800 at the short edge. We initialize a new track if the object has detection confidence higher than 0.0001.  

\parsection{Test: BDD100K}
The BDD100K contains 200 videos (40k) for validation and 400 videos (80k) for testing. We use both the BDD100K validation set and the test set for evaluation. For the inference, we use the $(1296, 720)$ image scale. We use the best performed model in the validation set which is saved at 2 epoch. We initialize a new track if the detection confidence is higher than 0.7.